\documentclass{article}

\PassOptionsToPackage{numbers, compress}{natbib}
\usepackage[preprint]{neurips_2026}

\usepackage[utf8]{inputenc}
\usepackage[T1]{fontenc}
\usepackage{microtype}

\usepackage{amsmath,amssymb,amsfonts,amsthm}
\usepackage{mathtools}
\usepackage{nicefrac}
\usepackage{bm}              
\usepackage{pgfplots}
\pgfplotsset{compat=1.18}

\newtheorem{theorem}{Theorem}
\newtheorem{lemma}[theorem]{Lemma}
\newtheorem{corollary}[theorem]{Corollary}

\newtheorem{remark}{Remark}

\usepackage{algorithm}
\usepackage{algpseudocodex}

\usepackage{booktabs}         
\usepackage{multirow}
\usepackage{array}

\usepackage{graphicx}
\usepackage{subcaption}       
\usepackage{tikz}
\usetikzlibrary{arrows.meta, positioning, calc, shapes.geometric, fit}

\usepackage{xcolor}
\definecolor{cuphblue}{HTML}{2C5282}
\definecolor{accentgreen}{HTML}{2F855A}
\definecolor{cmpred}{HTML}{C53030}
\definecolor{crispercolor}{HTML}{B83280}    
\definecolor{gudhigray}{HTML}{718096}        

\usepackage{hyperref}
\hypersetup{
  colorlinks=true,
  linkcolor=cuphblue,
  citecolor=cuphblue,
  urlcolor=cuphblue,
  bookmarksnumbered=true,
}
\usepackage{url}
\usepackage{cleveref}         

\usepackage{enumitem}         
\usepackage{wrapfig}          
\usepackage{etoolbox}         

\newcommand{\cuph}{\textnormal{\textsc{CuPH}}}

\title{CuPH: Parallel Cubical Persistent Homology via Level-Synchronous Component Tracking}

\author{
Guangyu Meng$^{1}$ \qquad
Pengfei Gu$^{2}$ \qquad
Erin Wolf Chambers$^{1}$ \qquad
Danny Z. Chen$^{1}$\\[0.5em]
$^{1}$Dept.\ of Computer Science and Engineering, University of Notre Dame\\
$^{2}$Dept.\ of Computer Science, The University of Texas Rio Grande Valley\\[0.3em]
{\tt\small \{gmeng, echambe2, dchen\}@nd.edu, \quad pengfei.gu01@utrgv.edu}
}

\begin{document}
\maketitle

\begin{abstract}
Persistent homology (PH) is a mathematically principled multi-scale topological descriptor of shape and connectivity in data. It is widely applied in image classification, segmentation, and shape analysis across medical and aerial imaging, providing structural information that pixel-wise deep learning may overlook. Standard cubical PH decomposes an image into cubical cells (vertices, edges, and squares) and reduces them sequentially in filtration order. This sequential reduction has $\Theta(n)$ parallel depth for an $n$-pixel image, limiting within-image GPU parallelism and constraining its use in topology-aware deep learning. We introduce CuPH, a GPU-based parallel algorithm for computing cubical PH on 2D images that operates directly on pixels rather than cubical cells. CuPH buckets pixels by filtration values, activates them with a level-synchronous sweep, and tracks connected components with a lock-free union-find method. We prove that CuPH produces persistence diagrams identical to standard cubical PH on integer inputs, with a stability bound for floating-point inputs. CuPH reduces parallel depth from $\Theta(n)$ to $O(L\log n)$ for an image with $L$ intensity levels. We implement the level sweep as a single CUDA kernel, eliminating cross-level kernel-launch overhead. Empirically, CuPH is $18.6$--$99.6\times$ faster than the existing GPU baseline (CMP). As a drop-in PH backend, CuPH accelerates topology-aware classification ($4.2$--$16.7\times$) and segmentation ($8.3$--$44\times$) while preserving accuracy. It also enables computing full-image PH losses on high-resolution images, improving segmentation Dice by $2.2$--$4.6$ points over patch-local PH baselines and accelerating training by $25$--$31\times$.
\end{abstract}

\section{Introduction}
\label{sec:introduction}

In image segmentation and classification, preserving topological structures (e.g., connected components and loops) is often as important as pixel-wise features. For example, a vessel-segmentation prediction may well overlap with ground-truth pixels but break a single capillary into multiple segments~\citep{shit2021cldice}; a road-extraction model may correctly label the road pixels but sever a highway at an intersection~\citep{mosinska2018beyond}; an organ-segmentation prediction may achieve high pixel agreement but introduce a phantom hole in the organ wall~\citep{clough2020topological}. Such errors violate the \emph{topology} of the target, which standard objectives such as cross-entropy and Dice~\citep{milletari2016vnet} do not explicitly constrain. Persistent homology (PH)~\citep{edelsbrunner2002topological,zomorodian2005computing}, a tool from computational topology, fills this gap by providing a principled measure of these structural properties. PH has motivated a growing body of topology-aware learning methods that use persistence-based features or losses to penalize broken vessels, fragmented road networks, leaky membranes, and spurious components~\citep{hu2019topology,stucki2023betti,lux2025topograph,peng2024phg,korkmaz2025cumperlay}.

\begin{figure*}[t]
\centering
\begin{subfigure}[t]{0.58\textwidth}
\centering
\caption{Pipeline comparison.}
\label{fig:pipeline}
\vspace{2pt}
\resizebox{\textwidth}{!}{
\begin{tikzpicture}[
  pxbox/.style={draw=gray!60, fill=gray!5, rounded corners=2pt,
    minimum width=1.4cm, minimum height=0.95cm, align=center,
    font=\small, line width=0.5pt},
  slowbox/.style={draw=cmpred!70, fill=cmpred!4, rounded corners=2pt,
    minimum width=2.3cm, minimum height=0.95cm, align=center,
    font=\small, line width=0.6pt},
  fastbox/.style={draw=accentgreen!70, fill=accentgreen!4, rounded corners=2pt,
    minimum width=2.6cm, minimum height=0.95cm, align=center,
    font=\small, line width=0.6pt},
  pdbox/.style={draw=gray!60, fill=gray!5, rounded corners=2pt,
    minimum width=1.4cm, minimum height=0.95cm, align=center,
    font=\small, line width=0.5pt},
  arr/.style={-{Stealth[length=3.5pt]}, gray!55, line width=0.7pt},
  slowtag/.style={font=\sffamily\bfseries\scriptsize,
    fill=cmpred!75!black, text=white,
    rounded corners=1.5pt, inner sep=1.5pt},
  fasttag/.style={font=\sffamily\bfseries\scriptsize,
    fill=accentgreen!65!black, text=white,
    rounded corners=1.5pt, inner sep=1.5pt},
]
\def\toprow{1.2}
\def\botrow{-1.2}
\node[pxbox] (img) at (0.0, 0.0) {Image\\$n$ pixels};
\node[pdbox] (pd) at (11.0, 0.0) {Persistence\\diagram};
\node[font=\small\bfseries, anchor=west, text=cmpred!70!black]
  at (1.2, \toprow+1.25) {Existing methods {\small(GUDHI, Cripser, CMP)}};
\node[slowbox] (e-build) at (2.5, \toprow)
  {Construct full\\complex\\{\footnotesize${\sim}4n$ cubical cells}};
\node[slowbox] (e-sort) at (5.0, \toprow)
  {Sort by\\filtration};
\node[slowbox] (e-uf) at (8.0, \toprow)
  {Sequential\\reduction\\{\footnotesize UF or matrix reduction}};
\node[slowtag, anchor=south] at (e-uf.north) {$\Theta(n)$ depth};
\draw[arr] (img.north east) to[out=25, in=180] (e-build.west);
\draw[arr] (e-build) -- (e-sort);
\draw[arr] (e-sort) -- (e-uf);
\draw[arr] (e-uf.east) to[out=0, in=155] (pd.north west);
\node[font=\small\bfseries, anchor=west, text=accentgreen!50!black]
  at (1.2, \botrow+1.25) {\textsc{CuPH} {\small(ours)}};
\node[fastbox] (c-bucket) at (3.5, \botrow)
  {Level\\Bucketing};
\node[fastbox] (c-sweep) at (7.5, \botrow)
  {Level Sweeping\\+ Component Tracking};
\node[fasttag, anchor=south] at (c-sweep.north) {$O(L\log n)$ depth};
\draw[arr] (img.south east) to[out=-25, in=180] (c-bucket.west);
\draw[arr] (c-bucket) -- (c-sweep);
\draw[arr] (c-sweep.east) to[out=0, in=-155] (pd.south west);
\node[font=\footnotesize\itshape, text=accentgreen!50!black, anchor=center]
  at (5.5, 0.0) {works directly on pixels in parallel};
\end{tikzpicture}}
\end{subfigure}%
\hfill
\begin{subfigure}[t]{0.40\textwidth}
\centering
\caption{Throughput comparison.}
\label{fig:batch}
\vspace{2pt}
\resizebox{\textwidth}{!}{
\begin{tikzpicture}
\begin{semilogyaxis}[
  width=\textwidth,
  height=0.7\textwidth,
  xmode=log,
  log basis x=2,
  xlabel={Batch size},
  ylabel={Throughput (images/s)},
  xlabel style={
    at={(ticklabel cs:1.0)},
    anchor=west,
    xshift=4pt,
    yshift=8pt,
    font=\small,
  },
  xmin=12, xmax=5500,
  ymin=8, ymax=55000,
  xtick={16,64,256,1024,4096},
  xticklabels={16,64,256,{1,024},{4,096}},
  xticklabel style={font=\footnotesize},
  ytick={10,100,1000,10000},
  yticklabels={10,100,{1,000},{10,000}},
  yticklabel style={font=\footnotesize},
  ylabel style={font=\small},
  grid=major,
  major grid style={gray!22, line width=0.4pt},
  yminorgrids=true,
  minor grid style={gray!10, line width=0.2pt},
  legend columns=2,
  legend style={
    at={(0.5, -0.20)},
    anchor=north,
    draw=none,
    font=\footnotesize,
    column sep=8pt,
    row sep=2pt,
  },
  clip=false,
]
\draw[cuphblue!70, dashed, line width=2pt]
  (axis cs:16, 21.8) -- (axis cs:16, 1452);
\draw[cuphblue!70, dashed, line width=2pt]
  (axis cs:128, 97.7) -- (axis cs:128, 9729);
\draw[cuphblue!70, dashed, line width=2pt]
  (axis cs:4096, 1441) -- (axis cs:4096, 26794);
\addplot[color=cuphblue, mark=*, mark size=2pt, line width=1.8pt]
  coordinates {
    (16,1452) (32,2792) (64,5232) (128,9729) (256,15443)
    (512,22366) (1024,22963) (2048,26164) (4096,26794)
  };
\addlegendentry{\textbf{CuPH} (GPU)}
\addplot[color=cmpred, mark=square*, mark size=2pt, line width=1.8pt]
  coordinates {
    (16,21.8) (32,31.6) (64,56.2) (128,97.7) (256,162.3)
    (512,286.4) (1024,513.1) (2048,931.5) (4096,1441)
  };
\addlegendentry{CMP (GPU)}
\addplot[color=crispercolor, dashed, line width=1.2pt,
  dash pattern=on 6pt off 3pt, forget plot]
  coordinates {(12,458.5) (5500,458.5)};
\addlegendimage{color=crispercolor, dashed, line width=1.2pt}
\addlegendentry{Cripser (CPU)}
\addplot[color=gudhigray, dashed, line width=1.2pt,
  dash pattern=on 6pt off 3pt, forget plot]
  coordinates {(12,97.5) (5500,97.5)};
\addlegendimage{color=gudhigray, dashed, line width=1.2pt}
\addlegendentry{GUDHI (CPU)}
\node[font=\footnotesize\bfseries, text=cuphblue, anchor=west]
  at (axis cs:17, 178) {66.6$\times$};
\node[font=\footnotesize\bfseries, text=cuphblue, anchor=west]
  at (axis cs:135, 1700) {99.6$\times$\,{\scriptsize(peak)}};
\node[font=\footnotesize\bfseries, text=cuphblue, anchor=east]
  at (axis cs:3900, 6217) {18.6$\times$};
\end{semilogyaxis}
\end{tikzpicture}}
\end{subfigure}
\caption{%
  \textbf{(a)} Existing cubical PH methods construct $\sim$$4n$ cubical cells (vertices, edges, and squares) and reduce them sequentially via union-find or matrix reduction. \cuph{} operates directly on pixels in parallel.
  \textbf{(b)} \cuph{} achieves $18.6$--$99.6\times$ speedup over the GPU baseline (CMP) across batch sizes on the ISIC\,2018 dataset~\cite{isic2018}; CPU methods plateau below $500$~img/s.
}
\label{fig:overview}
\end{figure*}
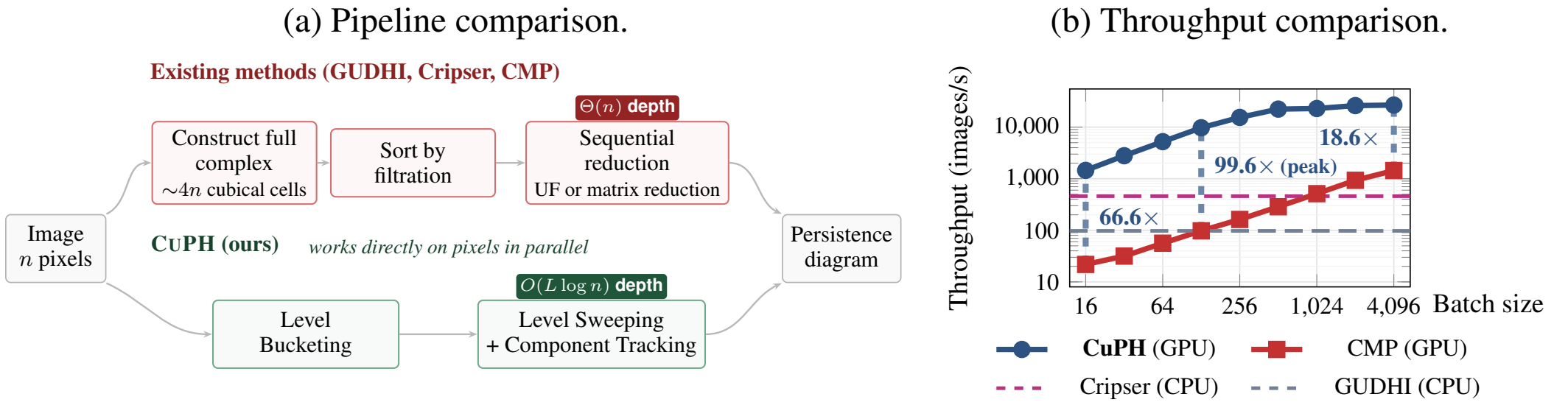

Despite the theoretical advantages, applying PH in deep learning (DL) remains computationally expensive. Standard cubical PH operates at the cell level (\Cref{fig:overview}a): it constructs all $\sim$$4n$ cubical cells (vertices, edges, and squares) of an $n$-pixel image, imposes a filtration order, and reduces them sequentially via union-find or matrix reduction methods~\citep{maria2014gudhi,kaji2020cubicalripser}. This serial reduction has $\Theta(n)$ parallel depth, which limits within-image parallelism on GPU. Consequently, even on $224^2$-size images, the fastest CPU algorithm, Cripser~\citep{kaji2020cubicalripser}, reaches only $458$ images/s, and the GPU baseline CMP~\citep{korkmaz2025cumperlay} requires batch $\geq 1{,}024$ to match Cripser, reaching $56$ images/s at batch $64$ and $1{,}441$ images/s only at batch $4{,}096$ (\Cref{fig:overview}b). Thus, researchers rely on three workarounds. \emph{Offline pre-computation} extracts PH features first and freezes them as model input~\citep{peng2024phg}; this avoids training-time computation but cannot adapt to a learned representation. \emph{Patch-local PH} restricts topological signals to small patches~\citep{hu2019topology,stucki2023betti} (typically sizes $65^2$ or $128^2$) in images that are often $1000^2$ or larger. This fits within the backend's per-image cost limits but cannot represent a global road network, membrane sheet, or organ boundary whose topological support extends beyond a single patch. \emph{Component-graph topology} replaces PH by a graph-theoretic proxy~\citep{lux2025topograph}, gaining speed but losing PH's theoretical guarantee that zero loss implies matching Betti numbers~\citep{hu2019topology}. A fast exact PH backend will change not only the runtime, but also the class of topology-aware objectives that DL models can use.

To address this computational bottleneck, we propose \cuph{}, a GPU algorithm for cubical PH. Our observation is that the cost of standard cubical PH is not inherent: it stems from cell-level formulations that do not exploit the bounded level structure of image filtrations. Specifically, for an image, the filtration parameter value is bounded and discrete, with $L = 256$ intensity levels for an $8$-bit image and $L \ll n$ in general. \cuph{} avoids constructing the full cubical complex (\Cref{fig:overview}a). It comprises two stages supported by a component-tracking primitive: \emph{Level Bucketing} groups pixels by intensity; \emph{Level Sweeping} processes levels sequentially, activating pixels at each level in parallel; \emph{Component Tracking} maintains connected components and emits persistence pairs under the Elder Rule~\citep{edelsbrunner2002topological}. \cuph{} thus reduces the parallel depth from $\Theta(n)$ to $O(L\log n)$, and achieves $18.6$--$99.6\times$ speedup over the GPU baseline (CMP) across batch sizes (\Cref{fig:overview}b).

Our contributions are as follows.
(i) \textbf{Algorithm:} A new level-synchronous GPU cubical PH algorithm for 2D images; 
\cuph{} operates directly on pixels, avoids explicit construction of the full cubical complex, and is implemented as a batched CUDA algorithm that can serve as a drop-in backend for existing topology-aware DL pipelines.
(ii) \textbf{Theory:} We prove that \cuph{} produces persistence diagrams (PDs) identical to standard cubical PH on integer inputs, with a stability bound for floating-point inputs. The algorithm has $O(n\log n + L)$ total work and $O(L\log n)$ parallel depth for an $n$-pixel image with $L$ intensity levels.
(iii) \textbf{Empirical results:} We evaluate \cuph{} as a standalone PH algorithm and as a drop-in backend for topology-aware classification, segmentation, and PH-based losses. \cuph{} is $18.6$--$99.6\times$ faster than the existing GPU baseline (CMP) across batch sizes. As a drop-in backend, it accelerates topology-aware classification by $4.2$--$16.7\times$ and segmentation by $8.3$--$44\times$ while preserving accuracy. It also enables computing full-image PH losses on high-resolution images, improving segmentation Dice by $2.2$--$4.6$ points over patch-local PH baselines (whose losses are computed only on local patches) and accelerating training by $25$--$31\times$.

\section{Preliminaries}
\label{sec:preliminaries}
Persistent homology (PH)~\citep{edelsbrunner2002topological,zomorodian2005computing} quantifies the topology of an image by tracking how connected components ($H_0$) and loops ($H_1$) appear and disappear across a filtration. Let $X: \Omega \to \{0, 1, \ldots, L-1\}$ denote an image on the pixel grid $\Omega$ of size $H \times W$, with $n = HW$ pixels, and $L$ intensity levels (e.g., $L = 256$ for an 8-bit image). For a pixel $p \in \Omega$, $X(p)$ denotes its intensity value, which serves as the filtration value at $p$. The cubical complex on $\Omega$ consists of vertices, edges, and squares induced by pixel adjacency (4- or 8-connectivity). Each cell $\sigma$ takes the filtration value $\hat{X}(\sigma) = \max_{p \in \mathrm{vertices}(\sigma)} X(p)$ (its largest constituent pixel intensity). At threshold $k$, all cells with $\hat{X}(\sigma) \leq k$ form the \emph{sublevel complex} $K_{\leq k}$, equivalently the cells whose constituent pixels all satisfy $X(p) \leq k$. As $k$ increases from $0$ to $L-1$, these complexes form a nested filtration $K_{\leq 0} \subseteq K_{\leq 1} \subseteq \cdots \subseteq K_{\leq L-1}$.

A connected component is \emph{born} at a threshold value where it first appears, and \emph{dies} when an edge merges it with another component. The \emph{Elder Rule}~\citep{edelsbrunner2010computational} resolves merges: the component with a lower birth value survives. In this work, ties are broken by lexicographic order on $(\mathrm{birth}, \mathrm{pixel\_index})$. Each death of a component yields a \emph{persistence pair} $(b, d)$ (with birth $b$ and death $d$). Pairs with $b = d$ lie on the diagram diagonal, called \emph{diagonal pairs}. Components that never die give \emph{essential pairs} $(b, +\infty)$. All these pairs form the \emph{persistence diagram} $\mathrm{PD}(X)$. We follow the standard convention that PDs identify diagonal pairs $(k, k)$ with the diagonal~\citep{edelsbrunner2010computational}; under this convention, diagonal pairs do not affect any diagram comparison, and thus we ignore them. PDs are often compared using the \emph{bottleneck} distance $d_B$~\citep{cohen2007stability} or \emph{$p$-Wasserstein} distance $W_p$~\citep{skraba2020wasserstein}, both used for optimal matchings between diagrams; these distances underpin the stability guarantees (discussed in \Cref{sec:analysis}).

\subsection{The Standard Cubical PH Algorithm}
\label{sec:standard_pipeline}

The standard cubical PH algorithm~\citep{kaji2020cubicalripser, maria2014gudhi} follows a three-step pipeline (\Cref{fig:overview}a, top): construct the full cubical complex with $\Theta(n)$ cells in $O(n)$ time, sort the cells by filtration value in $O(n \log n)$ time, and reduce them sequentially by edge-based union-find for $H_0$ and matrix reduction for higher $H$ dimensions. For $H_0$, union-find under the Elder Rule runs in $O(n \log n)$ amortized time~\citep{tarjan1984worst}. For $H_1$, the standard approach is boundary-matrix reduction~\citep{edelsbrunner2010computational}, with $O(n^3)$ worst-case complexity. For 2D images, this cubic cost can be avoided: \emph{Alexander duality}~\citep{edelsbrunner2012alexander, bleile2022persistent} recasts $H_1$ as a separate $H_0$ problem on a padded dual image, allowing union-find to handle both dimensions. CuPH adopts this duality-based reduction approach (\Cref{sec:method}).

The sequential nature of the edge-based union-find leads to an $\Theta(n)$ parallel depth, limiting GPU acceleration. The best-known GPU approach CMP~\citep{korkmaz2025cumperlay} batches across images but cannot parallelize within a single image. This bottleneck has constrained fast computation of PH-based features and differentiable topology losses in DL methods~\citep{hu2019topology, stucki2023betti, peng2024phg}. In particular, persistence-based segmentation losses commonly compute the topological term on small random patches for predictions rather than on the full image~\citep{hu2019topology, stucki2023betti}, capturing only local rather than global topological patterns.
\begin{figure}[t]
\centering
\includegraphics[width=0.90\textwidth, height=0.15\textheight, keepaspectratio=false]{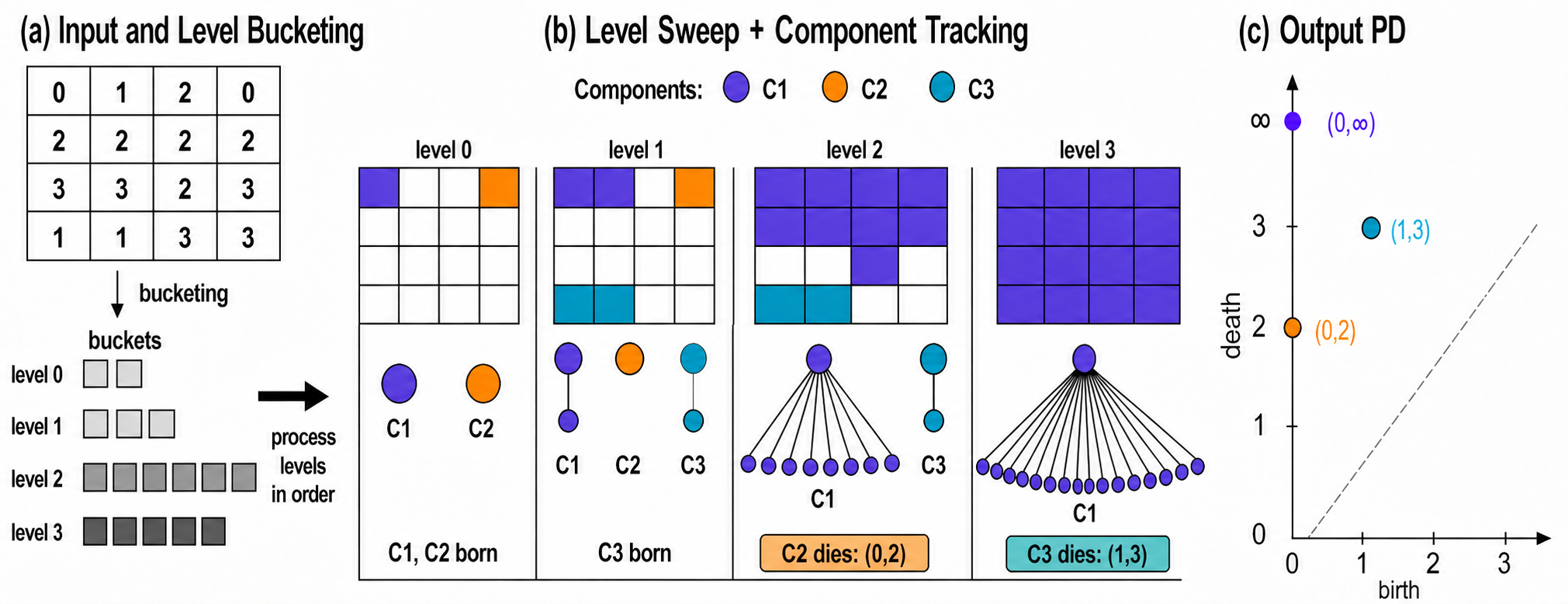}
\caption{Illustrating the CuPH pipeline on a $4 \times 4$ image with $L=4$ intensity levels.
\textbf{(a)}~Pixels are partitioned into buckets by intensity value.
\textbf{(b)}~Levels are processed in order: at level $0$, components $C_1$ and $C_2$ are born; at level $1$, $C_3$ is born, and one pixel joins $C_1$; at level $2$, $C_2$ merges into $C_1$, emitting a non-diagonal pair $(0,2)$; at level $3$, $C_3$ merges into $C_1$, emitting $(1,3)$.
\textbf{(c)}~The resulting persistence diagram, with $C_1$ giving an essential pair $(0,+\infty)$.}
\label{fig:cuph_pipeline}
\end{figure}

\section{Method}
\label{sec:method}

CuPH replaces the standard filtration sorting with level bucketing, followed by a level-synchronous parallel sweep that processes each level's pixels concurrently. We present the algorithm in \Cref{sec:pipeline}, the theoretical analysis in \Cref{sec:analysis}, and the GPU implementation in \Cref{sec:gpu-impl}.

\subsection{The CuPH Pipeline}
\label{sec:pipeline}

CuPH consists of two stages, \emph{Level Bucketing} and \emph{Level Sweeping}, supported by a \emph{Component Tracking} primitive invoked during the sweeping (\Cref{fig:cuph_pipeline}). Level Bucketing groups pixels by filtration value. Level Sweeping iterates over the levels in increasing order, activating each level's pixels in parallel and merging neighboring components. Component Tracking maintains a lock-free union-find forest, applying the Elder Rule~\citep{edelsbrunner2002topological} and emitting persistence pairs when components die. \Cref{alg:cuph} summarizes the procedure.

\textbf{Level Bucketing.} CuPH partitions the $n$ pixels of image $X$ into $L$ buckets by filtration value (pixel intensity), where $\textit{bucket}[k]$ contains all pixels $p$ with $X(p) = k$ (see \Cref{fig:cuph_pipeline}(a)). This step uses the standard GPU counting-sort primitive, which combines parallel histogram, prefix sum~\citep{harris2007parallel}, and scatter~\citep{merrill2010revisiting}, running in $O(n+L)$ work and $O(\log n)$ depth.

\textbf{Level Sweeping.} Level Sweeping consists of an outer loop across levels (sequential) and an inner loop within each level (parallel). The outer loop runs from level $k = 0$ to $L-1$ in order, because the Elder Rule at level $k$ depends on which components are active and on their birth values, both determined by earlier levels. Although the outer loop is not parallel, it shortens the sequential dependency chain from $\Theta(n)$ cells in the standard algorithm to $L$ levels. The inner loop is fully parallel: all pixels activated at level $k$ share the same intensity value, and processing them in any order yields the same persistence diagram (\Cref{thm:equivalence}).

At each level $k$, the inner loop has two phases separated by a barrier. The \emph{Activate} phase writes label and birth value for every pixel $p \in \textit{bucket}[k]$ in parallel. The \emph{Connect} phase then processes, in parallel, every neighboring pair $(p, q)$ with $p \in \textit{bucket}[k]$ and $q$ in the 4-connected neighborhood $\mathcal{N}(p)$ of $p$ that is already active (either from an earlier level or just activated at level $k$), invoking $\textsc{Merge}(p, q, k)$ (the details in Component Tracking). The barrier between the two phases ensures that pixels activated at level $k$ can all participate in within-level merges. \Cref{fig:cuph_pipeline}(b) illustrates this at level $k=1$: components $C_1$ and $C_2$ were born at level $0$, and three pixels in $\textit{bucket}[1]$ are activated (each as a singleton component with birth value $1$). Connect then merges them with their active neighbors: one pixel is absorbed into $C_1$ via an earlier-level neighbor; the other two merge with each other to form $C_3$. In each merge at level $k=1$, the component being absorbed is born at level $1$, giving a persistence pair $(1,1)$ that is diagonal and should be omitted. No pair is emitted at level $1$.

\textbf{Component Tracking.} CuPH maintains the active connected-component partition with a {\em union-find forest}. Each active pixel $p$ stores a parent pointer $\textit{label}[p]$, and each root $r$ (a pixel with $\textit{label}[r] = r$) stores its birth value $\textit{birth}[r]$ for the component it represents. The forest is updated via a $\textsc{Merge}$ function invoked by Connect; concurrent updates use lock-free atomic compare-and-swap (CAS).

$\textsc{Merge}(p, q, k)$ follows the standard lock-free union-find process, with the Elder Rule providing the linking criterion and successful merges emitting persistence pairs. Specifically, $\textsc{Merge}$ first runs $\textsc{Find}$ on $p$ and $q$ to locate their current roots. With the two roots found by $\textsc{Find}$, $\textsc{Merge}$ applies the Elder Rule using the lexicographic key $\kappa(r) = (\textit{birth}[r], r)$: the root with the smaller key becomes the \emph{survivor root}, and the other becomes the \emph{dying root}. The merge then links the dying root to the survivor root via an atomic CAS operation. If the CAS succeeds, the dying root is no longer a root and the merge is completed. If it fails, then another thread has linked this root in the meantime. The failed thread then reruns $\textsc{Find}$ on $p$ and $q$. Since the roots may have changed, the thread retries the merge, or returns if both $p$ and $q$ are found to already be in the same component.

A successful CAS triggers pair emission unless the merge would produce a diagonal pair (i.e., $(k,k)$). If the dying root $u$ was born before the current level (i.e., $\textit{birth}[u] < k$), CuPH emits a persistence pair $(\textit{birth}[u], k)$; if $\textit{birth}[u] = k$, the pair $(k,k)$ is diagonal and is omitted. \Cref{fig:cuph_pipeline}(b) illustrates a non-diagonal pair emission at level $k=2$, where level-$2$ activations cause $C_1$ and $C_2$ (both born at level $0$) to merge. The Elder Rule's tie-breaker (root indices) selects $C_1$ as the survivor root; a successful CAS links $C_2$ to $C_1$, $C_2$ dies, and CuPH emits the persistence pair $(\textit{birth}[u], k) = (0, 2)$.

\begin{algorithm}[bt!]
\caption{CuPH.}
\label{alg:cuph}
\footnotesize
\begin{algorithmic}[1]
\Require Image $X:\Omega\to\{0,\ldots,L-1\}$ with $n=|\Omega|$ pixels, 4-connected neighborhood $\mathcal{N}(p)$
\Ensure Persistence diagram $\mathrm{PD}(X)$
\State $\mathrm{PD}\gets\emptyset$
\State Partition $\Omega$ into $\textit{bucket}[0],\ldots,\textit{bucket}[L-1]$ by filtration value $X(p)$ \Comment{Level Bucketing}
\State \textbf{parallel for} $p\in\Omega$: $\textit{label}[p]\gets\bot$ \Comment{$\bot$: inactive}
\For{$k=0$ to $L-1$} \Comment{Level Sweeping}
    \State \textbf{parallel for} $p\in\textit{bucket}[k]$: $\textit{label}[p]\gets p$, $\textit{birth}[p]\gets k$ \Comment{Activate}
    \State \textbf{barrier}
    \State \textbf{parallel for} $p\in\textit{bucket}[k]$, $q\in\mathcal{N}(p)$ with $\textit{label}[q]\neq\bot$: $\textsc{Merge}(p,q,k)$ \Comment{Connect}
    \State \textbf{barrier}
\EndFor
\State append $(\min_p X(p),+\infty)$ to $\mathrm{PD}$ \Comment{essential pairs}
\State \Return $\mathrm{PD}$
\\
\Function{Merge}{$p,q,k$}
\Repeat
    \State $r_1\gets\textsc{Find}(p)$;\quad $r_2\gets\textsc{Find}(q)$ \Comment{locate current roots}
    \If{$r_1=r_2$} \State \Return \EndIf \Comment{already in the same component}
    \State $\kappa(r)\gets(\textit{birth}[r],r)$;\quad let $v$ (\emph{survivor root}), $u$ (\emph{dying root}) be the roots with smaller and larger $\kappa$, respectively 
    \State $\textit{old}\gets\textsc{AtomicCAS}(\textit{label}[u],u,v)$ \Comment{link $u$ to $v$}
    \If{$\textit{old}=u$} \Comment{CAS succeeded}
        \If{$\textit{birth}[u]<k$}
            \State atomically append $(\textit{birth}[u],k)$ to $\mathrm{PD}$ \Comment{emit a non-diagonal pair}
        \EndIf
        \State \Return
    \EndIf
\Until{false}
\EndFunction
\end{algorithmic}
\end{algorithm}

\subsection{Theoretical Analysis}
\label{sec:analysis}

We establish three guarantees for CuPH on an $n$-pixel, $L$-level image $X$: equivalence with the standard cubical PH (\Cref{thm:equivalence}), $O(n \log n + L)$ total work and $O(L \log n)$ parallel depth (\Cref{thm:work,thm:parallel-depth}), and stability bounds (Corollaries~\ref{cor:stability} and~\ref{cor:quantization-stability}). Let $\mathrm{PD}_{\mathrm{CuPH}}^{H_q}(X)$ denote CuPH's output diagram and $\mathrm{PD}_V^{H_q}(X)$ denote the standard cubical diagram, for $H_q \in \{H_0, H_1\}$.
All bounds are analyzed for one image. For a batch $B$ of images, the work scales by $B$ and the parallel depth is unchanged.

\textbf{Equivalence.} Both the standard cubical PH algorithm and CuPH process the same neighboring pixel pairs $(p, q)$ at level $\max(X(p), X(q))$: the standard algorithm via an explicit edge, while CuPH via Connect with both endpoints $p$ and $q$ being active. The two therefore feed the same merges to union-find at each level but differ in their merge order (sequential vs. parallel). \Cref{thm:equivalence} shows that this order difference does not affect the output.

\begin{theorem}[Equivalence]
\label{thm:equivalence}
For an image $X: \Omega \to \{0, \ldots, L-1\}$,
\[
\mathrm{PD}_{\mathrm{CuPH}}^{H_q}(X) = \mathrm{PD}_V^{H_q}(X), \quad \text{for } H_q \in \{H_0, H_1\}.
\]
\end{theorem}
\begin{proof}[Proof Sketch]
We prove the equivalence by induction on the filtration level $k$, with three key properties at each level: \emph{identical inputs} (both algorithms feed the same set of merges to union-find), \emph{order-invariance} (all sequential merge orders produce the same persistence diagram), and \emph{linearization} (CuPH's concurrent process is equivalent to some valid sequential merge order). See Appendix~\ref{app:equivalence} for the full proof.
\end{proof}

\begin{theorem}[Work complexity]
\label{thm:work}
Assume that the atomic CAS operation in each $\textsc{Merge}$ conducts amortized $O(1)$ retries. CuPH performs $O(n \log n + L)$ total work on image $X$.
\end{theorem}

\begin{proof}[Proof Sketch]
In CuPH, Level Bucketing performs $O(n + L)$ work via histogram, prefix sum, and scatter. Level Sweeping processes $O(n)$ neighboring pairs (each pixel has $O(1)$ neighbors); each pair invokes $\textsc{Merge}$, which under the assumption performs amortized $O(1)$ retries, each with $O(\log n)$ amortized cost from $\textsc{Find}$~\citep{tarjan1984worst}. Thus, the total work on $X$ is $O(n + L) + O(n \log n) = O(n \log n + L)$. The full proof is in Appendix~\ref{app:complexity}.
\end{proof}

\begin{theorem}[Parallel depth]
\label{thm:parallel-depth}
Consider an ideal shared-memory model with atomic compare-and-swap (CAS). Assume that $L \leq n$, the number $P$ of processors satisfies $P \geq c \max_k |\textit{bucket}[k]|$, where $c$ is the neighborhood size (e.g., $c = 4$ for 4-connectivity), the CAS operation in each $\textsc{Merge}$ conducts amortized $O(1)$ retries, and any $\textsc{Find}$ path in the union-find forest is of a length $O(\log n)$ under the maintained union-find forest invariant. Then CuPH has a parallel depth of $O(L \log n)$.
\end{theorem}

\begin{proof}[Proof Sketch]
With $L \leq n$, Level Bucketing has $O(\log n)$ parallel depth via the standard work-efficient parallel counting-sort primitive. Level Sweeping performs $L$ sequential iterations (the Elder Rule at level $k$ depends on the earlier levels); within each iteration, the Activate phase has $O(1)$ depth, and the Connect phase has $O(\log n)$ depth from $\textsc{Find}$ via path halving, with amortized $O(1)$ retries conducted by CAS adding only a constant factor overhead. In total, $O(\log n) + L \cdot O(\log n) = O(L \log n)$ parallel depth. The full proof is in Appendix~\ref{app:complexity}.
\end{proof}

\textbf{Stability.} CuPH's output diagrams inherit standard cubical PH stability via \Cref{thm:equivalence}.

\begin{corollary}[Stability]
\label{cor:stability}
For any two discrete-valued images $X, Y: \Omega \to \{0, \ldots, L-1\}$ on a 2D pixel grid $\Omega$, with $H_q \in \{H_0, H_1\}$, bottleneck distance $d_B$, and Wasserstein distance $W_p$,
\begin{align*}
d_B\bigl(\mathrm{PD}_{\mathrm{CuPH}}^{H_q}(X),\, \mathrm{PD}_{\mathrm{CuPH}}^{H_q}(Y)\bigr) &\leq \|X - Y\|_\infty, \\
W_p\bigl(\mathrm{PD}_{\mathrm{CuPH}}^{H_q}(X),\, \mathrm{PD}_{\mathrm{CuPH}}^{H_q}(Y)\bigr) &\leq 9\, \|X - Y\|_p, \quad p \in [1, \infty).
\end{align*}
\end{corollary}
By \Cref{thm:equivalence}, the bounds follow from the bottleneck~\citep{cohen2007stability} and Wasserstein~\citep[Theorem~5.1]{skraba2020wasserstein} stability.

\textbf{Quantization Stability.} In practice, image inputs often contain continuous-values. The next corollary bounds the deviation of CuPH's diagram from the PD of the original continuous-valued image.

\begin{corollary}[Quantization stability]
\label{cor:quantization-stability}
Let the value range of a function $f: \Omega \to \mathbb{R}$ be $[a, b]$ (i.e., $a \leq f(x) \leq b$), $X$ be the uniform $L$-level quantization of $f$ with step size $\Delta = (b-a)/L$ (hence, $X(p) = \min(\lfloor (f(p) - a)/\Delta \rfloor, L-1)$), and $\tilde{X}(p) = a + (X(p) + 1/2)\Delta$ denote its bin-midpoint dequantization, so that $\|f - \tilde{X}\|_\infty \leq \Delta/2$. Let $\overline{\mathrm{PD}}_{\mathrm{CuPH}}^{H_q}(X)$ denote CuPH's output diagram with each value $k$ rescaled to $a + (k + 1/2)\Delta$. Then for $H_q \in \{H_0, H_1\}$,
\begin{align*}
d_B\bigl(\overline{\mathrm{PD}}_{\mathrm{CuPH}}^{H_q}(X),\, \mathrm{PD}_V^{H_q}(f)\bigr) &\leq \Delta/2, \\
W_p\bigl(\overline{\mathrm{PD}}_{\mathrm{CuPH}}^{H_q}(X),\, \mathrm{PD}_V^{H_q}(f)\bigr) &\leq 9\, n^{1/p}\, \Delta/2, \quad p \in [1, \infty).
\end{align*}
\end{corollary}
By \Cref{thm:equivalence}, $\overline{\mathrm{PD}}_{\mathrm{CuPH}}^{H_q}(X) = \mathrm{PD}_V^{H_q}(\tilde{X})$. The bottleneck bound follows from $\|f - \tilde{X}\|_\infty \leq \Delta/2$ and the bottleneck stability~\citep{cohen2007stability}. The Wasserstein bound follows from~\citet[Theorem~5.1]{skraba2020wasserstein} with $\|f - \tilde{X}\|_p \leq n^{1/p}\Delta/2$. \Cref{tab:correctness} in the Appendix reports that empirically, $W_2$ is approximately $80\times$ tighter than this worst-case Wasserstein bound.

\subsection{GPU Implementation}
\label{sec:gpu-impl}
CuPH's level-synchronous structure poses two GPU implementation challenges. (a) \emph{Inter-level synchronization}: level $k+1$ cannot start until all Activate and Connect operations at level $k$ finish, and per-level kernel launch would incur overhead proportional to $L$. (b) \emph{Intra-level concurrency}: within each level, multiple $\textsc{Merge}$ calls may simultaneously update the union-find forest, requiring atomic operations to prevent lost updates. We address these with two design choices: a single-kernel level sweep and a lock-free union-find. We 
describe batched execution and the $H_1$ pipeline below.

\textbf{Single-kernel Level Sweep.} To address inter-level synchronization, CuPH runs the entire level sweep within one kernel, with each image assigned to a single CUDA block. The block iterates over the levels in ascending order; at each level, threads cooperatively execute the Activate phase (writing labels and birth values for pixels in the level's bucket), hit a \texttt{\_\_syncthreads()} barrier, and then execute the Connect phase (visiting active neighbors and calling $\textsc{Merge}$). A second \texttt{\_\_syncthreads()} at the end of each level ensures that all level-$k$ merges complete before level-$(k+1)$ activations start. Because the entire sweep runs in a single block, this level-synchronous iteration uses inexpensive block-local barriers in place of repeated kernel launches. Buckets larger than the block's thread count are processed via the standard CUDA strided-loop pattern.

\textbf{Lock-free Union-Find.} To address intra-level concurrency, CuPH implements the union-find forest lock-free. Per-pixel labels and birth values reside in global memory. $\textsc{Find}$ uses path halving for efficiency~\citep{tarjan1984worst}. $\textsc{Merge}$ links the dying root to the survivor root via \texttt{atomicCAS} on the dying root's label. Emitted persistence pairs are recorded via \texttt{atomicAdd} on the per-image pair counter. Forest validity follows from a strict-decrease invariant established in the proof of \Cref{thm:equivalence}: each successful CAS attaches a dying root to a parent with a strictly smaller lexicographic key $\kappa(r) = (\textit{birth}[r], r)$; thus $\kappa$ strictly decreases along every parent pointer regardless of CAS interleaving, precluding cycles.

\textbf{Batch Parallelism and Variants.} A pre-allocated memory pool gives each image its own private workspace. Independent images therefore run in independent CUDA blocks without contention, enabling batched processing across the GPU. The $H_1$ pipeline reuses the same kernel after constructing the padded Alexander-dual image~\citep{bleile2022persistent}; superlevel filtrations are obtained by reversing intensity levels and remapping the resulting birth--death pairs. Full kernel pseudocode, memory layout, and additional implementation details are in Appendix~\ref{app:gpu}.

\section{Experiments}
\label{sec:experiments}
We evaluate \cuph{} in two distinct roles. First, as a standalone GPU PH method, we benchmark its throughput and scalability against GUDHI, Cripser, and CMP across image resolutions and batch sizes (\Cref{tab:throughput}). Second, as a drop-in backend for topology-aware learning, we put \cuph{} into classification pipelines (\Cref{tab:cls_main}) and into two segmentation settings: topology-aware architectures (\Cref{tab:seg_accuracy}), and topology-aware losses (\Cref{tab:topoloss}).

\noindent
\textbf{Experimental Setup.}
\emph{Datasets.} For classification, we use CBIS-DDSM~\citep{cbis_ddsm}, Prostate Cancer (Prostate)~\citep{lawson2019prostate}, ISIC\,2018 Task~3 (T3)~\citep{isic2018}, and APTOS\,2019~\citep{aptos2019}, following PHG-Net~\citep{peng2024phg} and CMP~\citep{korkmaz2025cumperlay}. For segmentation under the topology-aware architecture setting, we use ISIC\,2018 Task~1 (T1)~\citep{isic2018}, CREMI~\citep{cremi2016} (resized to $512^2$), and Synapse~\citep{landman2015miccai} (2D axial slices). For topology-aware loss experiments, we follow Topograph~\citep{lux2025topograph} on Massachusetts Roads (Roads)~\citep{mnih2013}, Platelet~\citep{guay2021platelet}, and CREMI (native $1250^2$). Dataset details are summarized in \Cref{tab:datasets} (Appendix~\ref{app:datasets}).

\emph{PH Baselines.} We compare four cubical PH methods: \textbf{GUDHI}~\citep{maria2014gudhi} and \textbf{Cripser}~\citep{kaji2020cubicalripser} (both CPU), \textbf{CMP}~\citep{korkmaz2025cumperlay} (the only published GPU cubical PH), and \textbf{\cuph{}}. \cuph{} discretizes float-valued images into $L=2^{10}$ levels, which matches the practical precision of float segmentation probability maps and yields empirical $W_2$ well below the bound of \Cref{cor:quantization-stability} (Appendix~\ref{app:correctness}). To obtain $H_1$, Cripser, CMP, and \cuph{} use cubical Alexander duality~\citep{bleile2022persistent}, and GUDHI applies matrix reduction.

\emph{Methods.} \textbf{Classification} (SwinV2-B backbone~\citep{liu2022swin}): backbone alone (no PH); {PHG-Net}~\citep{peng2024phg} (original offline-precomputed version and online \cuph{} variants); {TopoSwin-SP/MP}~\citep{korkmaz2025cumperlay} (single- and multi-parameter persistence, with CMP and \cuph{} backends). \textbf{Segmentation} uses a SwinV2-B encoder with a U-Net style decoder~\citep{ronneberger2015unet}; TopoSwin-SP/MP apply persistence as decoder-skip losses, following~\citep{korkmaz2025cumperlay}. \textbf{Topology-aware losses}: HuTopo~\citep{hu2019topology} and BettiMatching~\citep{stucki2023betti} (PH-based, with \cuph{} backend); Topograph~\citep{lux2025topograph} (non-PH component-graph reference).

\emph{Training.} All methods trained with AdamW~\citep{loshchilov2019decoupled} ($\mathrm{lr}{=}10^{-4}$, weight decay $0.01$) in fp32 with cosine schedule, using the original papers' hyperparameters when available. \cuph{}-backed variants inherit settings from their CMP-backed counterparts. GPU experiments use a single NVIDIA H100; CPU on AMD EPYC 7543. Full hyperparameters in Appendix~\ref{app:experimental-details}. Code will be released upon acceptance.

\emph{Metrics.} Accuracy and Dice are mean$_{\pm\mathrm{std}}$ over $5$ seeds; throughput (img/s) is mean over $5$ runs. Classification reports top-1 accuracy; segmentation reports Dice and mIoU (ISIC, CREMI) or Dice and HD95~mm (Synapse); topology-aware losses additionally report clDice~\citep{shit2021cldice}.

\begin{table}[t]
\caption{%
  Throughput comparison on ISIC 2018. (a)~Throughput (img/s) at varying resolutions (batch size = $64$). (b)~Throughput, memory (MB), and speedup at varying batch sizes (resolution = $224^2$).
}
\label{tab:throughput}
\centering
\footnotesize
\setlength{\tabcolsep}{3pt}
\begin{minipage}[t]{0.42\textwidth}
\centering
\subcaption{Resolution scaling}
\label{tab:resolution_scaling}
\begin{tabular}{@{}lrrrrr@{}}
\toprule
\textbf{Method}
  & $64^2$ & $128^2$ & $224^2$
  & $512^2$ & $1024^2$ \\
\midrule
GUDHI       & $1{,}232$  & $269.8$    & $97.5$    & $9.4$    & $1.8$ \\
Cripser     & $5{,}258$  & $1{,}495$  & $458.5$   & $38.7$   & $9.1$ \\
CMP         & $1{,}349$  & $268.4$    & $56.2$    & $7.7$    & $1.3$ \\
\cuph{}     & $11{,}319$ & $8{,}453$  & $5{,}232$ & $805.2$  & $208.4$ \\
\midrule
\cuph{}/CMP & $8.4\times$ & $31.5\times$ & $93.1\times$
            & $104.6\times$  & $160.3\times$ \\
\bottomrule
\end{tabular}
\end{minipage}
\hfill
\begin{minipage}[t]{0.54\textwidth}
\centering
\subcaption{Batch scaling}
\label{tab:batch_sensitivity}
\begin{tabular}{@{}rrrrrr@{}}
\toprule
\textbf{Batch size}
  & \multicolumn{2}{c}{\textbf{\cuph{}}}
  & \multicolumn{2}{c}{\textbf{CMP}}
  & \textbf{Speedup} \\
\cmidrule(lr){2-3} \cmidrule(lr){4-5}
  & img/s & MB & img/s & MB & \\
\midrule
$16$   & $1{,}452$   & $170$      & $21.8$    & $298$      & $66.6\times$ \\
$32$   & $2{,}792$   & $190$      & $31.6$    & $496$      & $88.4\times$ \\
$64$   & $5{,}232$   & $272$      & $56.2$    & $994$      & $93.1\times$ \\
$128$  & $9{,}729$   & $428$      & $97.7$    & $1{,}408$  & $99.6\times$ \\
$256$  & $15{,}443$  & $748$      & $162.3$   & $2{,}716$  & $95.2\times$ \\
$1024$ & $22{,}963$  & $2{,}898$  & $513.1$   & $10{,}398$ & $44.8\times$ \\
$4096$ & $26{,}794$  & $11{,}292$ & $1{,}441$ & $41{,}264$ & $18.6\times$ \\
\bottomrule
\end{tabular}
\end{minipage}
\vspace{-10pt}
\end{table}

\noindent
\textbf{Standalone PH Benchmarking.}
\emph{Resolution Scaling} (\Cref{tab:resolution_scaling}).
On ISIC with batch $64$, \cuph{} achieves $5{,}232$ img/s at resolution $224^2$, a $93.1\times$ speedup over CMP and $11.4\times$ over Cripser. It sustains the highest throughput across all resolutions tested, from $11{,}319$ img/s at $64^2$ down to $208.4$ img/s at $1024^2$. Throughput declines with increasing image size since per-image work scales with $n$ while \cuph{}'s per-image thread block keeps the within-image parallelism constant. In contrast, CMP launches one thread per image, and thus its per-image work is fully sequential and its throughput stays near GUDHI's despite running on GPU. Hence, as image size grows, CMP's throughput declines steeply while \cuph{}'s within-image parallelism keeps the decline slow. The speedup ratio of \cuph{} over CMP grows accordingly, reaching $160.3\times$ at $1024^2$.

\emph{Batch Scaling} (\Cref{tab:batch_sensitivity}).
At image size $224^2$ on ISIC 2018, \cuph{} scales from $1{,}452$ img/s at batch size $16$ to $26{,}794$ img/s at batch size $4{,}096$, leading at every batch size. The \cuph{}-over-CMP speedup peaks at $99.6\times$ at batch size $128$ and narrows at both extremes: at small batch sizes, the GPU is under-utilized for both methods, while at large batch sizes, CMP saturates its only source of parallelism (cross-images) and begins to close the gap. Even so, CMP's reliance on cross-image parallelism alone shows that batch size cannot compensate for limited within-image parallelism. \cuph{}'s memory advantage shows a similar small-batch effect: \cuph{} uses $3.3$--$3.7\times$ less GPU memory than CMP at batch size $\geq 64$, narrowing to $1.8$--$2.6\times$ at batch sizes $16$ and $32$ where \cuph{}'s level-bucketing 
buffers contribute a larger fixed share of its memory footprint.

\emph{Ablation.} Removing Level Bucketing slows \cuph{} by $4.6\times$, restricting Level Sweeping to a single thread by $21.0\times$, and removing both jointly by $425.4\times$. This joint slowdown exceeds the product of the two individual slowdowns ($4.6 \times 21.0 \approx 97\times$) since single-threaded scanning over an unsorted filtration incurs additional sequential overhead. Replacing lock-free Component Tracking by a global lock slows down the kernel by $19.8\times$. Full ablation is given in Appendix~\ref{app:ablation}.
Beyond throughput, \cuph{} produces PDs identical to standard cubical PH on integer images; for floating-point images, the quantization error satisfies \Cref{cor:quantization-stability} and its empirical verification is given in Appendix~\ref{app:correctness}.

\begin{table}[t]
\caption{%
  Classification accuracy (Acc, \%) and training throughput (img/s) at batch size $32$. 
}
\label{tab:cls_main}
\centering
\footnotesize
\setlength{\tabcolsep}{3pt}
\begin{tabular}{@{}l cc cc cc cc@{}}
\toprule
& \multicolumn{2}{c}{\textbf{ISIC 2018}}
& \multicolumn{2}{c}{\textbf{CBIS-DDSM}}
& \multicolumn{2}{c}{\textbf{Prostate}}
& \multicolumn{2}{c}{\textbf{APTOS}} \\
\cmidrule(lr){2-3}\cmidrule(lr){4-5}\cmidrule(lr){6-7}\cmidrule(lr){8-9}
\textbf{Method} & Acc & img/s & Acc & img/s & Acc & img/s & Acc & img/s \\
\midrule
SwinV2-B
  & $88.1_{\pm 1.9}$ & $206.7$
  & $69.4_{\pm 0.3}$ & $206.1$
  & $96.0_{\pm 0.2}$ & $208.5$
  & $82.4_{\pm 1.7}$ & $55.2$  \\
\midrule
PHG-Net
  & $88.7_{\pm 0.9}$ & $204.0$
  & $71.8_{\pm 0.6}$ & $202.9$
  & $96.9_{\pm 0.6}$ & $202.7$
  & $84.1_{\pm 1.7}$ & $54.9$  \\
PHG-Net (\cuph{})
  & $89.5_{\pm 0.9}$ & $183.9$
  & $71.9_{\pm 1.1}$ & $183.4$
  & $96.6_{\pm 0.3}$ & $183.7$
  & $85.0_{\pm 1.2}$ & $49.5$  \\
\midrule
TopoSwin-SP
  & $88.2_{\pm 0.5}$ & $16.4$
  & $73.8_{\pm 0.2}$ & $16.8$
  & $96.6_{\pm 0.4}$ & $17.9$
  & $86.1_{\pm 1.5}$ & $2.9$   \\
TopoSwin-SP (\cuph{})
  & $89.3_{\pm 0.4}$ & $116.9$
  & $74.1_{\pm 0.3}$ & $111.1$
  & $96.4_{\pm 0.5}$ & $118.5$
  & $86.3_{\pm 1.4}$ & $28.8$ \\
\midrule
TopoSwin-MP
  & $89.8_{\pm 1.0}$ & $12.1$
  & $74.3_{\pm 0.1}$ & $12.4$
  & $96.9_{\pm 0.2}$ & $11.9$
  & $83.9_{\pm 1.4}$ & $0.6$  \\
TopoSwin-MP (\cuph{})
  & $90.3_{\pm 0.6}$ & $52.6$
  & $74.5_{\pm 0.2}$ & $52.6$
  & $97.0_{\pm 0.3}$ & $53.3$
  & $84.1_{\pm 1.1}$ & $10.0$ \\
\bottomrule
\end{tabular}
\vspace{-10pt}
\end{table}

\noindent
\textbf{Topology-aware Classification.}
We evaluate \cuph{} using three topology-aware classification pipelines on four medical image benchmarks (\Cref{tab:cls_main}). PHG-Net~\citep{peng2024phg} originally precomputes PH features offline; the \cuph{} variant computes them online during training. TopoSwin-SP and TopoSwin-MP~\citep{korkmaz2025cumperlay} originally use CMP, which we replace by \cuph{} as a drop-in backend.

Based on the results in \Cref{tab:cls_main}, we make three observations. First, replacing CMP by \cuph{} matches or improves the accuracy across all benchmarks; TopoSwin-MP (\cuph{}) achieves the best accuracy on ISIC 2018, CBIS-DDSM, and Prostate; on APTOS, TopoSwin-SP (\cuph{}) leads. Second, replacing PHG-Net's offline-precomputed PH features by \cuph{}'s online-computed features adds $10$--$11\%$ training overhead while preserving the accuracy within $\pm 1\%$, eliminating the need for offline PH precomputation. Third, \cuph{} accelerates TopoSwin training by $4.2$--$7.1\times$ on the three $224^2$ datasets and $9.9$--$16.7\times$ on the $512^2$ APTOS, consistent with \cuph{}'s growing advantage at higher resolutions (\Cref{tab:resolution_scaling}). Minor accuracy differences between PH backends are within typical run-to-run variances; \cuph{}'s PDs match the standard cubical PH exactly on integer images (Appendix~\ref{app:correctness}). Peak GPU memory and AUC are reported in Table~\ref{tab:efficiency-full} of the Appendix.

\begin{table}[t]
\caption{%
  Segmentation results with training throughput (img/s). The metrics follow each benchmark's standard convention: Dice and mIoU for ISIC\,2018 (T1) and CREMI; Dice and HD95 for Synapse.
}
\label{tab:seg_accuracy}
\centering
\footnotesize
\setlength{\tabcolsep}{3pt}
\begin{tabular}{@{}l ccc ccc ccc@{}}
\toprule
& \multicolumn{3}{c}{\textbf{ISIC 2018 (T1)}}
& \multicolumn{3}{c}{\textbf{CREMI}}
& \multicolumn{3}{c}{\textbf{Synapse}} \\
\cmidrule(lr){2-4} \cmidrule(lr){5-7} \cmidrule(lr){8-10}
\textbf{Method}
  & Dice$\uparrow$ & mIoU$\uparrow$ & img/s
  & Dice$\uparrow$ & mIoU$\uparrow$ & img/s
  & Dice$\uparrow$ & HD95$\downarrow$ & img/s \\
\midrule
Baseline (no PH)
  & $89.8_{\pm 1.1}$ & $79.2_{\pm 1.3}$ & $212.3$
  & $83.8_{\pm 1.1}$ & $74.7_{\pm 0.6}$ & $38.6$
  & $82.3_{\pm 0.8}$ & $17.6_{\pm 0.3}$ & $236.3$ \\
\midrule
TopoSwin-SP
  & $90.9_{\pm 1.4}$ & $80.4_{\pm 1.5}$ & $12.9$
  & $84.4_{\pm 0.9}$ & $75.1_{\pm 0.7}$ & $1.4$
  & $82.6_{\pm 1.0}$ & $16.6_{\pm 0.6}$ & $13.5$ \\
TopoSwin-SP (\cuph{})
  & $91.2_{\pm 1.2}$ & $80.5_{\pm 1.1}$ & $171.5$
  & $84.3_{\pm 0.8}$ & $75.2_{\pm 0.5}$ & $17.8$
  & $82.9_{\pm 0.6}$ & $16.5_{\pm 0.2}$ & $189.1$ \\
\midrule
TopoSwin-MP
  & $91.5_{\pm 1.3}$ & $81.3_{\pm 1.4}$ & $9.8$
  & $84.1_{\pm 1.3}$ & $74.9_{\pm 0.6}$ & $0.3$
  & $82.4_{\pm 1.4}$ & $16.2_{\pm 0.5}$ & $9.4$ \\
TopoSwin-MP (\cuph{})
  & $91.6_{\pm 1.1}$ & $81.2_{\pm 0.9}$ & $81.2$
  & $84.3_{\pm 0.6}$ & $75.0_{\pm 0.7}$ & $13.1$
  & $82.5_{\pm 0.8}$ & $16.4_{\pm 0.4}$ & $88.6$ \\
\bottomrule
\end{tabular}
\vspace{-10pt}
\end{table}

\noindent
\textbf{Topology-aware Architectures for Segmentation.}
We evaluate \cuph{} as a drop-in PH backend for TopoSwin-SP/TopoSwin-MP
on three benchmarks of varying resolutions (\Cref{tab:seg_accuracy}). The architectures and topology loss formulation follow the Experimental Setup; we replace CMP by \cuph{}.

\Cref{tab:seg_accuracy} reveals two patterns. (i) \emph{Segmentation performance is preserved.} In all $12$ comparisons between the original and \cuph{} variants (TopoSwin-SP vs.\ TopoSwin-SP~(\cuph{}), TopoSwin-MP vs.\ TopoSwin-MP~(\cuph{}), each on three benchmarks and two metrics), the largest mean difference is $0.3$ points, within cross-seed variance. (ii) \emph{Throughput becomes practical.} At size $224^2$, \cuph{} accelerates TopoSwin-SP by $13.3$--$14.0\times$ over CMP and TopoSwin-MP by $8.3$--$9.4\times$; at size $512^2$ on CREMI, the multi-parameter gap widens to $44\times$ for TopoSwin-MP, taking it from impractical ($0.3$ img/s) to practical ($13.1$ img/s) and enabling multi-parameter topology-aware segmentation at this resolution.

\noindent
\textbf{Topology-aware Losses for Segmentation.}
We evaluate \cuph{} as a drop-in PH backend for HuTopo~\citep{hu2019topology} and BettiMatching~\citep{stucki2023betti} on the three topology-aware loss datasets, with Topograph~\citep{lux2025topograph} as a non-PH reference. The original HuTopo and BettiMatching compute PH on $128^2$ patches due to CPU backend cost; the \cuph{} variants compute PH on the full probability map in every training step.

\Cref{tab:topoloss} reports two findings. First, full-image PH supervision (enabled by \cuph{}'s computational throughput) improves both Dice ($+2.2$ to $+4.6$ points) and clDice ($+2.0$ to $+4.4$ points) over patch-local PH for both topology losses across all three datasets. Patch-local PH operates on $128^2$ regions rather than the full image, capturing local rather than global topology; this was a workaround for CPU backends~\citep{maria2014gudhi, kaji2020cubicalripser} where full-image computation was a bottleneck. The accuracy gains therefore reflect the move from patch-local to full-image topology supervision, made practical by \cuph{}, rather than a pure backend substitution. Second, \cuph{} accelerates training by $25$--$31\times$ over patch-local PH backends, with BettiMatching (\cuph{}) running $4$--$7\times$ faster than Topograph on every dataset, demonstrating that PH-based topology-aware training at full resolution is now practical.

\begin{table}[t]
\caption{%
  \cuph{} as a drop-in replacement for CPU PH backends in topology-aware segmentation losses, evaluated at full resolution (Dice and clDice in \%, training throughput in img/s, batch size $8$).
}
\label{tab:topoloss}
\centering
\footnotesize
\setlength{\tabcolsep}{3pt}
\begin{tabular}{@{}l ccc ccc ccc@{}}
\toprule
& \multicolumn{3}{c}{\textbf{CREMI}}
& \multicolumn{3}{c}{\textbf{Roads}}
& \multicolumn{3}{c}{\textbf{Platelet}} \\
\cmidrule(lr){2-4} \cmidrule(lr){5-7} \cmidrule(lr){8-10}
\textbf{Method}
  & Dice$\uparrow$ & clDice$\uparrow$ & img/s
  & Dice$\uparrow$ & clDice$\uparrow$ & img/s
  & Dice$\uparrow$ & clDice$\uparrow$ & img/s \\
\midrule
Topograph (no PH)
  & $90.2_{\pm 1.0}$ & $94.8_{\pm 0.8}$ & $2.51$
  & $81.1_{\pm 2.3}$ & $68.7_{\pm 1.9}$ & $1.42$
  & $70.4_{\pm 3.1}$ & $78.0_{\pm 2.7}$ & $5.47$ \\
\midrule
HuTopo
  & $90.1_{\pm 0.8}$ & $94.6_{\pm 0.6}$ & $0.19$
  & $80.7_{\pm 1.1}$ & $66.7_{\pm 1.5}$ & $0.08$
  & $68.2_{\pm 2.0}$ & $74.4_{\pm 1.8}$ & $0.44$ \\
HuTopo (\cuph{})
  & $94.7_{\pm 0.4}$ & $96.6_{\pm 0.5}$ & $5.54$
  & $84.8_{\pm 1.3}$ & $70.2_{\pm 1.2}$ & $2.46$
  & $71.3_{\pm 1.5}$ & $78.8_{\pm 1.4}$ & $11.01$ \\
\midrule
BettiMatching
  & $88.1_{\pm 0.7}$ & $91.3_{\pm 0.5}$ & $0.47$
  & $80.3_{\pm 0.9}$ & $65.9_{\pm 0.8}$ & $0.34$
  & $69.8_{\pm 1.3}$ & $76.1_{\pm 1.0}$ & $0.78$ \\
BettiMatching (\cuph{})
  & $90.3_{\pm 0.5}$ & $93.5_{\pm 1.0}$ & $14.38$
  & $82.6_{\pm 0.8}$ & $68.3_{\pm 0.6}$ & $10.02$
  & $72.3_{\pm 1.1}$ & $79.9_{\pm 0.9}$ & $24.22$ \\
\bottomrule
\end{tabular}
\vspace{-10pt}
\end{table}

\section{Conclusions and Discussion}
\label{sec:conclusion}

We presented \cuph{}, a new GPU algorithm for computing cubical PH that operates directly on pixels rather than cubical cells, reducing parallel depth from $\Theta(n)$ to $O(L\log n)$. \cuph{} produces persistence diagrams identical to standard cubical PH on integer images, with a stability bound for floating-point images, and is $18.6$--$99.6\times$ faster than the existing GPU baseline.

\textbf{Extending to Higher Dimensions.} \cuph{} is currently formulated for 2D images, on which Alexander duality reduces $H_1$ to an $H_0$ problem on a padded dual image. In 3D images, Alexander duality relates $H_0$ to $H_2$, but $H_1$ in the middle dimension is self-dual and still requires matrix reduction. Extending \cuph{} to 3D thus requires a parallel approach for middle-dimensional homology (for example, through parallel column reduction or matching-based formulations).

\textbf{Beyond Uniform Quantization.} For floating-point images, \cuph{} relies on uniform $L$-level quantization (\Cref{cor:quantization-stability}). Adaptive or topology-aware quantization could tighten this bound or reduce $L$ with similar accuracy, particularly for non-uniformly distributed inputs such as sigmoid model outputs where most mass concentrates near $0$ and $1$.

\bibliographystyle{abbrvnat}
\bibliography{references}

\appendix
\newpage

\section{Proof of Theorem~\ref{thm:equivalence}}
\label{app:equivalence}

We prove that CuPH and the standard cubical PH algorithm under the V-construction produce identical non-diagonal persistence diagrams as multisets, for $H_q \in \{H_0, H_1\}$. The proof works at the algorithmic level on Algorithm~\ref{alg:cuph}; the GPU implementation in \Cref{sec:gpu-impl} is a faithful refinement and inherits correctness.

The $H_0$ case proceeds by induction on the filtration level $k$, with three steps at each level: \emph{identical inputs} (\Cref{sub:identical-inputs}), \emph{order-invariance} (\Cref{sub:order-invariance}), and \emph{linearization} (\Cref{sub:linearization}). The $H_1$ case reduces to $H_0$ via cubical Alexander duality (\Cref{sub:proof-h1}).

\paragraph{Setup.} Let $X: \Omega \to \{0, \ldots, L-1\}$ be an image on the pixel grid $\Omega$ with $n = HW$ pixels. Let $K$ be the V-construction cubical complex on $\Omega$: pixels are 0-cells, 1-cells join 4-adjacent pixels, 2-cells fill each $2 \times 2$ block, and $\hat{X}(\sigma) = \max_{p \in \mathrm{vertices}(\sigma)} X(p)$. We write $\mathrm{PD}^{>0}$ for the non-diagonal persistence diagram. For each activated pixel $p$, $\mathrm{birth}[p]$ denotes its birth level and $\kappa(p) = (\mathrm{birth}[p], p)$ its lexicographic key. The Elder Rule selects the root with smaller $\kappa$ as the survivor; the larger as the dying root.

\subsection{Step 1: Identical inputs}
\label{sub:identical-inputs}

Both algorithms process the same set of merges at each level $k$ with the same Elder-Rule semantics.

\begin{lemma}[Edge filtration value]
\label{lem:edge-value}
For every 1-cell $e \in K$ with endpoint pixels $p, q$, $\hat{X}(e) = \max(X(p), X(q))$.
\end{lemma}

\begin{proof}
The 0-cell faces of $e$ are $p$ and $q$, so $\hat{X}(e) = \max(X(p), X(q))$ by definition.
\end{proof}

Let $E_k = \{e \in K : \hat{X}(e) = k\}$ denote the set of 1-cells with filtration value $k$. By Lemma~\ref{lem:edge-value}, an edge $e = \{p,q\} \in E_k$ has both endpoints activated by level $k$ (since $\max(X(p), X(q)) = k$), so $e$ is processed at level $k$ in any V-construction filtration.

\begin{lemma}[Merge realization]
\label{lem:merge}
Fix a sequential Elder-Rule schedule on $E_k$ starting from the post-Activate state at level $k$. Let $e \in E_k$ have endpoints $p, q$ with current root births $b_p, b_q$. If $\mathrm{root}(p) = \mathrm{root}(q)$, no pair is emitted. Otherwise, the schedule emits $(\max(b_p, b_q), k)$, non-diagonal iff $b_p < k$ and $b_q < k$.
\end{lemma}

\begin{proof}
The Elder Rule kills the larger-$\kappa$ root, which has birth $\max(b_p, b_q)$. The pair $(\max(b_p, b_q), k)$ is non-diagonal iff $\max(b_p, b_q) < k$, equivalently both $b_p, b_q < k$.
\end{proof}

\subsection{Step 2: Order-invariance}
\label{sub:order-invariance}

\begin{lemma}[Within-level schedule independence]
\label{lem:partition}
Fix the union-find state after the Activate phase at level $k$, comprising pre-level-$k$ components together with newly activated level-$k$ singletons. For any sequential Elder-Rule schedule on $E_k$:
\begin{enumerate}[label=(\roman*), itemsep=0pt]
  \item the resulting partition of activated pixels is the same;
  \item the multiset of non-diagonal pairs emitted at level $k$ is the same.
\end{enumerate}
\end{lemma}

\begin{proof}
Let $R_1, \ldots, R_s$ be the components active after the Activate phase at level $k$ (including newly activated singletons), with keys $\kappa(R_i)$. Define the auxiliary multigraph $G_k$ on $\{R_1, \ldots, R_s\}$ with one edge per $e \in E_k$ joining the components containing its endpoints. Let $F_1, \ldots, F_t$ be the connected components of $G_k$.

\emph{Proof of (i).} Two components $R_i, R_j$ end up merged iff they lie in the same $F_l$, by standard connected-component / union-find equivalence. Hence the post-level-$k$ partition is $F_1, \ldots, F_t$, independent of edge order.

\emph{Proof of (ii).} Fix $F_l$ and let $R_l^* = \arg\min_{R_i \in F_l} \kappa(R_i)$ be its smallest-key component. We claim $R_l^*$ is the unique survivor in $F_l$ under any sequential Elder-Rule schedule. By the Elder Rule, every union kills the larger-$\kappa$ root. At every union involving $R_l^*$ (or its current super-component, whose root remains $R_l^*$), the merge partner's root lies in $F_l$ and has $\kappa$ no smaller than $\kappa(R_l^*)$; hence the partner dies and $R_l^*$ survives. Since $F_l$ is connected, all components in $F_l$ end up unified, leaving $R_l^*$ as the unique survivor.

The remaining $|F_l| - 1$ components $R_i \in F_l \setminus \{R_l^*\}$ all die at level $k$, each emitting the pair $(\mathrm{birth}[R_i], k)$, non-diagonal iff $\mathrm{birth}[R_i] < k$. The set of dying roots and their births is determined by $F_l$, so the non-diagonal contribution from $F_l$ is order-independent. Summing over $l$ gives the level-$k$ non-diagonal multiset.
\end{proof}

\subsection{Step 3: Linearization}
\label{sub:linearization}

We show that CuPH's concurrent execution at level $k$ corresponds to a valid sequential Elder-Rule schedule on $E_k$.

\begin{lemma}[Birth stability]
\label{lem:birth-stability}
For every pixel $p$, $\mathrm{birth}[p]$ is written exactly once at activation and never modified. Hence $\kappa(p)$ is constant after activation.
\end{lemma}

\begin{proof}
Algorithm~\ref{alg:cuph} writes $\mathrm{birth}[p]$ only in the Activate phase. $\textsc{Find}$ reads only $\mathrm{label}$, and $\textsc{Merge}$ writes only to $\mathrm{label}$.
\end{proof}

\begin{lemma}[CuPH linearization]
\label{lem:linearization}
Assume the forest invariant $\kappa(\mathrm{label}[x]) <_{\mathrm{lex}} \kappa(x)$ holds for every $x$ with $\mathrm{label}[x] \ne x$ at the start of CuPH's Connect phase at level $k$. Order the successful CAS operations of Connect by their atomic completion times, denoted $C_1, \ldots, C_m$. Then:
\begin{enumerate}[label=(\roman*), itemsep=0pt]
  \item \emph{Coverage.} Every $e \in E_k$ is offered to $\textsc{Merge}$ at least once during Connect.
  \item \emph{Forest invariant.} After each $C_i$ and after each path-halving write, the union-find is a valid disjoint-set forest with $\kappa(\mathrm{label}[x]) <_{\mathrm{lex}} \kappa(x)$ for all $x$ with $\mathrm{label}[x] \ne x$, and no cycles.
  \item \emph{Schedule realization.} The final partition and emitted pair multiset match those of some sequential Elder-Rule schedule on $E_k$.
\end{enumerate}
\end{lemma}

\begin{proof}
\emph{(i) Coverage.} Let $e \in E_k$ with endpoints $p, q$. By Lemma~\ref{lem:edge-value}, at least one endpoint (say $p$) lies in $\mathrm{bucket}[k]$. The Activate-Connect barrier ensures $q$ has $\mathrm{label}[q] \ne \bot$ when Connect begins. Hence $p$'s thread invokes $\textsc{Merge}(p, q, k)$.

\emph{(ii) Forest invariant.} We show every CAS commit and every path-halving write preserves the local invariant $\kappa(\mathrm{label}[x]) <_{\mathrm{lex}} \kappa(x)$, by induction over the linearization order of writes.

A successful $\textsc{CAS}(\mathrm{label}[u], u, v)$ requires $\mathrm{label}[u] = u$ at the linearization point, so $u$ is a root. The Elder Rule selected $\kappa(v) <_{\mathrm{lex}} \kappa(u)$ at issue time; by Lemma~\ref{lem:birth-stability}, this remains true at the CAS point. After the CAS, $\mathrm{label}[u] = v$ satisfies the invariant. A cycle would require a parent path $v \to \cdots \to u$ with $\kappa$ strictly decreasing, contradicting $\kappa(v) <_{\mathrm{lex}} \kappa(u)$.

A path-halving write replaces $\mathrm{label}[w]$ with a value $a$ that the writer observed at $\mathrm{label}[\mathrm{label}[w]]$. By the local invariant at read time, $\kappa(a) <_{\mathrm{lex}} \kappa(\mathrm{label}[w]) <_{\mathrm{lex}} \kappa(w)$, so $\kappa(a) <_{\mathrm{lex}} \kappa(w)$. By Lemma~\ref{lem:birth-stability}, this remains true at write time. Cycles are precluded by the same argument.

CAS writes occur only when $u$ is a root; once $u$ becomes non-root, subsequent CASes on $\mathrm{label}[u]$ fail. Path-halving writes target only non-roots. The two write types are temporally disjoint on the same node.

\emph{(iii) Schedule realization.} For each successful CAS $C_i = \textsc{CAS}(\mathrm{label}[u_i], u_i, v_i)$, the CAS premise $\mathrm{label}[u_i] = u_i$ at linearization ensures $u_i$ is still the root of one component at that point. Acyclicity (from (ii)) ensures the two roots $u_i, v_i$ belong to distinct components, since otherwise the new edge $u_i \to v_i$ would close a cycle. Hence $C_i$ effects an Elder-Rule union of two distinct components, emitting the pair prescribed by Lemma~\ref{lem:merge}. The sequence $C_1, \ldots, C_m$ realizes Elder-Rule unions on a subset $E_k' \subseteq E_k$; edges in $E_k \setminus E_k'$ have their endpoints already merged via prior CASes. Hence $E_k'$ produces the same partition and pair multiset as any sequential schedule on $E_k$. By Lemma~\ref{lem:partition}, this matches the standard algorithm's output at level $k$.
\end{proof}

\subsection{Inductive Combination for \texorpdfstring{$H_0$}{H0}}
\label{sub:proof-h0}

\begin{proof}[Proof of Theorem~\ref{thm:equivalence} for $H_0$]
We prove $\mathrm{PD}_{\mathrm{CuPH}}^{H_0,>0}(X) = \mathrm{PD}_V^{H_0,>0}(X)$ by induction on $k$. The inductive hypothesis is that, at the start of level $k$, both algorithms agree on (A) the activated-pixel partition, (B) the recorded births at roots, and (C) the non-diagonal pair multiset emitted so far, with (D) CuPH's forest satisfying the local invariant.

\emph{Base case ($k = 0$):} no pixels activated; (A)--(D) hold vacuously.

\emph{Inductive step:} assume (A)--(D) hold at the start of level $k$. Both algorithms activate level-$k$ pixels as singletons with $\mathrm{birth} = k$, preserving (A), (B), (D). The standard algorithm processes $E_k$ via a sequential Elder-Rule schedule. CuPH's Connect phase, by Lemma~\ref{lem:linearization}(iii) (whose forest-invariant assumption is provided by (D)), produces the same final partition and pair multiset as some sequential schedule on $E_k$. By Lemma~\ref{lem:partition}, all such schedules produce the same outcome. Hence (A)--(C) hold at the start of $k+1$, and (D) is preserved by Lemma~\ref{lem:linearization}(ii).

\emph{Essential class.} A 4-connected pixel grid has one $H_0$ essential class, born at $\min_p X(p)$. Both algorithms record this pair, completing Theorem~\ref{thm:equivalence} for $H_0$.
\end{proof}

\subsection{\texorpdfstring{$H_1$}{H1} via Alexander Duality}
\label{sub:proof-h1}

The $H_1$ case reduces to $H_0$ via cubical Alexander duality. Define the padded dual image $\tilde{X}^\vee$ on the enlarged grid $\tilde{\Omega} = \{0, \ldots, H+1\} \times \{0, \ldots, W+1\}$ by $\tilde{X}^\vee(p) = (L-1) - X(p)$ for $p \in \Omega$ and $\tilde{X}^\vee(p) = 0$ on the boundary frame. By cubical Alexander duality~\citep{bleile2022persistent,edelsbrunner2010computational}, $\mathrm{PD}_V^{H_1, >0}(X) = \Phi(\tilde{D}_0^{\mathrm{fin}, >0})$ as multisets, where $\tilde{D}_0$ is the $H_0$ persistence diagram of $\tilde{X}^\vee$ under the V-construction with 8-connectivity, $\tilde{D}_0^{\mathrm{fin}}$ excludes the unique essential class from the boundary frame, and $\Phi(b', d') = ((L-1) - d', (L-1) - b')$.

The $H_0$ equivalence proof above (Lemmas~\ref{lem:edge-value}--\ref{lem:linearization}) makes no use of the specific neighbor relation: Lemma~\ref{lem:edge-value} uses only the V-construction max rule, Lemma~\ref{lem:partition} depends only on the edge set $E_k$, and Lemma~\ref{lem:linearization}'s coverage argument requires only that each $e \in E_k$ have at least one endpoint in $\mathrm{bucket}[k]$. These hold for any neighbor relation, including 8-connectivity. Applied to $\tilde{X}^\vee$ with 8-connectivity, the proof gives $\mathrm{PD}_{\mathrm{CuPH}}^{H_0, >0}(\tilde{X}^\vee, \text{8-conn}) = \tilde{D}_0^{>0}$.

CuPH's $H_1$ pipeline (\Cref{sec:gpu-impl}) constructs $\tilde{X}^\vee$, runs the $H_0$ machinery with 8-connectivity, and applies $\Phi$ to each finite pair. Hence
\[
\mathrm{PD}_{\mathrm{CuPH}}^{H_1, >0}(X) = \Phi(\tilde{D}_0^{\mathrm{fin}, >0}) = \mathrm{PD}_V^{H_1, >0}(X),
\]
completing Theorem~\ref{thm:equivalence}. $\square$

\section{Proofs of Theorems~\ref{thm:work} and~\ref{thm:parallel-depth}}
\label{app:complexity}

\paragraph{Setup.} Let $n = HW$ be the number of pixels and $L$ the number of discrete intensity levels. \emph{Work} refers to the total number of operations across all threads; \emph{depth} refers to the longest sequential dependency chain.

\begin{lemma}[Path halving cost under elder-rule linking]
\label{lem:path-halving}
For a sequence of $m \geq n$ \textsc{Find} and \textsc{Union} operations
on a disjoint-set forest of $n$ elements using path halving with
arbitrary linking (in particular, elder-rule linking, where the union
step selects the smaller-key root as parent), the total work is
$O(m \log n)$, giving amortized cost $O(\log n)$ per operation.
\end{lemma}

\begin{proof}
Lemma~7 of~\citet{tarjan1984worst} establishes the bound for path
halving with naive (arbitrary-choice) linking: for $m \geq n$, the
total number of nodes traversed on find paths is at most
$(8m + 2n) \lceil \log_{1+m/n} n \rceil$, which is $O(m \log n)$.
The proof relies only on the path-halving invariant (each \textsc{Find}
halves the distance to the root along its path) and on the union step
taking constant time; it does not depend on which of the two roots
becomes the parent. Hence the bound applies verbatim to elder-rule
linking, which deterministically selects the smaller-key root as
parent. The classical $O(m\alpha(n))$ bound is unavailable here
because it requires union-by-rank or union-by-size~\citep{tarjan1984worst},
both of which conflict with elder rule's persistence-driven choice
of parent.
\end{proof}

\begin{lemma}[Sequential CAS does not retry]
\label{lem:cas-sequential}
In the sequential setting (one thread), the total number of \textsc{CAS} retries across all \textsc{Merge} calls is zero.
\end{lemma}

\begin{proof}
With one thread executing, no other thread modifies $\textit{label}[\textit{younger}]$ between the \textsc{Find} that produced $\textit{younger}$ and the subsequent \textsc{CAS}. The \textsc{CAS} therefore succeeds on the first attempt.
\end{proof}

\begin{lemma}[Sequential dependency across levels]
\label{lem:level-dependency}
The Connect phase at level $k$ requires the union-find state at the end of level $k-1$. The outer loop over levels in Algorithm~\ref{alg:cuph} therefore cannot be parallelized.
\end{lemma}

\begin{proof}
At level $k$, when pixel $p$ is in the Connect phase, it examines each 4-neighbor $q$ to determine whether $q$ is active and the current $\textsc{Find}$ representative of $q$. Both depend on level $k-1$'s processing being complete: if $q$ has $X(q) = k-1$, then $q$ became active during level $k-1$'s Activate phase and may have been merged into other components during level $k-1$'s Connect phase. Algorithm~\ref{alg:cuph} enforces this dependency through a barrier between consecutive levels: all threads complete level $k-1$'s Connect phase before any thread begins level $k$'s Activate phase, ensuring that level $k$ observes a consistent post-level-$(k-1)$ union-find state.
\end{proof}

\subsection{Proof of Theorem~\ref{thm:work}}

\begin{proof}
We decompose the work by stage.

\emph{Stage 1 (Level Bucketing).} The histogram pass performs $O(1)$ work per pixel, total $O(n)$. The exclusive prefix sum is $O(L)$. The scatter pass is $O(1)$ per pixel, total $O(n)$. Stage 1 total: $O(n + L)$.

\emph{Level Sweeping.} The outer loop iterates over $L$ levels. The Activate phase contributes $O(n)$ in total across all levels (each pixel activated once). The Connect phase contributes at most $4n$ merge attempts.

Each merge attempt performs two $\textsc{Find}$ operations and at most one $\textsc{CAS}$. By Lemma~\ref{lem:path-halving}, each $\textsc{Find}$ costs $O(\log n)$ amortized. By Lemma~\ref{lem:cas-sequential}, each $\textsc{CAS}$ succeeds on first attempt and contributes $O(1)$. Total: $4n \cdot O(\log n) = O(n \log n)$.

\emph{Total.} $O(n + L) + O(n \log n) = O(n \log n + L)$.
\end{proof}

\subsection{Proof of Theorem~\ref{thm:parallel-depth}}

\begin{proof}
We decompose the depth by stage in the combining-atomic PRAM model.

\emph{Stage 1 (Level Bucketing).} The histogram pass uses atomic increments to $L$ counters. In the combining-atomic model, $c$ concurrent atomic increments to one counter complete in $O(\log c) \leq O(\log n)$ depth. The exclusive prefix sum has $O(\log L)$ depth via parallel scan. The scatter pass has $O(\log n)$ depth from atomic contention. Stage 1 depth: $O(\log n + \log L) = O(\log n)$.

\emph{Level Sweeping.} By Lemma~\ref{lem:level-dependency}, the outer loop is sequential, contributing $L$ steps. Within each level: Activate has $O(1)$ depth (independent writes); the barrier has $O(\log P)$ depth; Connect processes pixels concurrently with up to 4 merge attempts each.

Each merge attempt has depth: $\textsc{Find}$ cost $O(\log n)$ (under path halving, the union-find tree height remains $O(\log n)$, so each $\textsc{Find}$ traverses at most $O(\log n)$ ancestors); $\textsc{CAS}$ attempt $O(1)$; constant amortized retry rounds, each adding $O(\log n)$. Per-merge depth: $O(\log n)$.

Per-level depth: $O(\log n)$. Total Level Sweeping depth: $L \cdot O(\log n) = O(L \log n)$.

\emph{Total.} $O(\log n) + O(L \log n) = O(L \log n)$.
\end{proof}

\begin{remark}
Relative to a comparison-sort-based total-order traversal (e.g., Cripser~\citep{kaji2020cubicalripser}) with $O(n \log n)$ sequential depth, the idealized CuPH level sweep reduces the dominant depth to $O(L \log n)$, a factor of $\Theta(n / L)$. For $L = 256$ and $n = 10^6$, this is approximately $4 \times 10^3$. The empirical wall-clock speedups, which incorporate constant-factor effects and finite-$P$ scheduling, are reported in Section~\ref{sec:experiments}.
\end{remark}

\section{GPU Implementation Details}
\label{app:gpu}

This section provides implementation details deferred from \Cref{sec:gpu-impl}.

\subsection{Kernel Decomposition}
\label{app:gpu-kernels}

CuPH is realized as a sequence of CUDA kernels following the two-stage structure of \cref{alg:cuph}. Stage~1 (Level Bucketing) is implemented by three distinct kernels: a histogram kernel that counts pixels per filtration level, an exclusive-prefix-sum kernel that converts these counts into per-level offsets, and a scatter kernel that writes each pixel's index into the corresponding per-level bucket. Stage~2 (Level Sweeping) is implemented by a single processing kernel that performs the Activate--Connect--Merge inner loop for each level. Output packing kernels then convert the internal pair buffers into output tensors and append the essential $H_0$ class corresponding to the global minimum; indexed variants additionally return the pixel indices responsible for each pair, used for gradient assignment in differentiable training.

\subsection{Memory Layout}
\label{app:gpu-memory}

For a batch of $B$ images of size $H \times W$ with $L$ filtration levels, CuPH allocates per-image arrays from a reusable device-memory pool: per-pixel \emph{label} and \emph{birth} arrays storing each pixel's union-find representative and birth filtration level; five per-level metadata arrays implementing the bucket structure (a histogram of pixel counts per level, exclusive-scan offsets into the bucket, per-level pixel counts, atomic scatter counters, and the bucketed pixel-index list itself); separate \emph{birth} and \emph{death} arrays sized to the maximum possible $H_0$ pairs per image; and per-image scalars (atomic pair counter, accepted-pair count, minimum filtration level). Indexed variants additionally allocate \emph{creator} and \emph{destroyer} arrays storing the pixel indices associated with each pair: the creator pixel is the dying root's representative at birth, and the destroyer pixel is the one whose activation triggered the merge. Independent images occupy disjoint regions of each array, allowing concurrent access from different blocks without cross-block synchronization.

\subsection{Processing Kernel: Activate--Connect--Merge}
\label{app:gpu-processing}

Each image is assigned to one CUDA block; threads in the block iterate over filtration levels in ascending order. At each nonempty level, the threads first execute the Activate phase, writing labels and birth values for all pixels in the level's bucket. A block-local barrier separates this from the Connect phase, in which threads visit each newly active pixel's neighbors and invoke the merge primitive whenever an active neighbor lies in an earlier or current level. The barrier ensures that all activations at level $k$ complete before any merges at level $k$ begin, preserving the level-synchronous invariant required by \cref{thm:equivalence}.

\subsection{Concurrent Merge Primitive}
\label{app:gpu-merge}

The merge primitive performs an elder-rule union of the components containing two given pixels. Representatives are located via $\textsc{Find}$ with path halving: at each step the search reads the current node's grandparent and writes it back as the new parent of the current node, halving the path length for subsequent queries (in contrast to full path compression, which sets every visited node's parent to the root in a second pass). Once the two representatives are found, the elder root (smaller lexicographic key $(\textit{birth}, \textit{index})$) is selected as the survivor; the dying root is attached to the survivor via \texttt{atomicCAS} on its self-loop label entry. On a successful CAS, non-diagonal pairs reserve an output slot via \texttt{atomicAdd} on the per-image pair counter and write the $(b, d)$ entry; indexed variants additionally record the creator and destroyer pixel indices. Failed CASes indicate that another thread has updated the dying root's parent in the meantime; the merge retries with refreshed representatives obtained from a new $\textsc{Find}$.

The number of CAS retries per merge is bounded by the number of concurrent merges affecting the same root, which is constant amortized in practice (consistent with the assumption in \cref{thm:parallel-depth}). Failed CASes emit no pair, so no spurious pairs are recorded. Forest validity follows from the monotone representative-key invariant proved in \cref{thm:equivalence} (Step~4 of the proof sketch): every successful CAS attaches a node to a parent with strictly smaller $\kappa = (\textit{birth}, \textit{index})$, precluding cycles regardless of CAS interleaving order.

\subsection{\texorpdfstring{$H_1$}{H1} via Padded Dual}
\label{app:gpu-h1}

The $H_1$ pipeline reuses the same kernel sequence as $H_0$ via cubical Alexander duality~\citep[Theorem~5]{bleile2022persistent}. The padded dual image $X^\vee$ is materialized in a small set of tensor operations preceding the kernel pipeline: pixel values in the interior are complemented as $X(p) \mapsto (L-1) - X(p)$, and a zero-valued boundary frame of width one is attached on all sides. These operations are dispatched as standard PyTorch elementwise GPU kernels, not as part of the CuPH processing kernel; their cost is negligible relative to the level sweep. The processing kernel then runs on $X^\vee$ with 8-connectivity (the topological dual of 4-connectivity), and the output packing applies the duality pair-transformation $\Phi(b', d') = ((L-1) - d', (L-1) - b')$ to each finite pair, excluding the essential class corresponding to the boundary frame. The result is the $H_1$ persistence diagram of the original image. Superlevel filtrations are handled analogously: the input is mapped via $X(p) \mapsto (L-1) - X(p)$ before the sweep, and pairs are transformed correspondingly during output.

\section{Experimental Details}
\label{app:experimental-details}
This appendix collects dataset preprocessing, PH algorithm implementation notes, per-dataset training hyperparameters, and the software and hardware stack used in \Cref{sec:experiments}.

\subsection{Dataset Details and Preprocessing}
\label{app:datasets}

\begin{table}[t]
\caption{%
Benchmark datasets.
}
\label{tab:datasets}
\centering
\small
\setlength{\tabcolsep}{4pt}
\begin{tabular}{@{}llcccl@{}}
\toprule
\textbf{Task} & \textbf{Dataset}
  & \textbf{Resolution}
  & \textbf{Classes}
  & \textbf{Train / Test}
  & \textbf{Domain} \\
\midrule
\multirow{4}{*}{Classification}
  & CBIS-DDSM       & $224^2$  & $2$ & $1{,}247$ / $624$     & Mammography    \\
  & Prostate Cancer & $224^2$  & $3$ & $4{,}145$ / $1{,}037$ & Histopathology \\
  & ISIC 2018 (T3)  & $224^2$  & $7$ & $9{,}013$ / $1{,}002$ & Dermatology    \\
  & APTOS 2019      & $512^2$  & $5$ & $3{,}296$ / $366$     & Ophthalmology  \\
\midrule
\multirow{3}{*}{Segmentation}
  & ISIC 2018 (T1)  & $224^2$           & $2$ & $2{,}698$ / $1{,}002$ & Dermatology         \\
  & CREMI           & $1250^2{}^{\dagger}$ & $2$ & $336$ / $39$          & Electron microscopy \\
  & Synapse         & $224^2$           & $9$ & $2{,}211$ / $1{,}568$ & Abdominal CT        \\
\midrule
\multirow{2}{*}{\shortstack[l]{Topology-aware\\loss}}
  & Roads    & $1500^2$ & $2$ & $1{,}122$ / $49$ & Aerial imagery      \\
  & Platelet & $800^2$  & $7$ & $50$ / $24$      & Electron microscopy \\
\bottomrule
\end{tabular}
\\[2pt]
\footnotesize${}^{\dagger}$~CREMI is used at native $1250^2$ for topology-aware loss experiments and resized to $512^2$ for the segmentation.
\end{table}

\paragraph{ISIC\,2018 (Task~3).}
The ISIC\,2018 Task~3 challenge~\citep{isic2018} contains $10{,}015$ dermoscopy images across $7$ diagnostic classes (melanoma, melanocytic nevus, basal cell carcinoma, actinic keratosis, benign keratosis, dermatofibroma, vascular lesion). The class distribution is highly imbalanced, with melanocytic nevus accounting for roughly two-thirds of the training set. We use the official challenge split, combining training and validation for training ($9{,}013$ images) and evaluating on the held-out test set ($1{,}002$ images). Images are resized to $224^2$ using bilinear interpolation and converted to 8-bit grayscale via the ITU-R BT.601 luma transform ($Y = 0.299R + 0.587G + 0.114B$) for PH computation; the RGB image is retained as input to the SwinV2-B backbone. Augmentations during training include random horizontal and vertical flips, random rotation ($\pm 15^{\circ}$), and color jitter (brightness/contrast/saturation $\pm 0.2$).

\paragraph{CBIS-DDSM.}
CBIS-DDSM~\citep{cbis_ddsm} provides curated mammography regions of interest with binary labels (benign vs.\ malignant). We use the provider's split, combining training and validation for training ($1{,}247$ images) and evaluating on the held-out test set ($624$ images). We extract the pre-cropped ROI patches, resize to $224^2$, and apply min-max normalization to $[0, 255]$ on the single-channel DICOM intensities. Augmentations include random horizontal flip and random rotation ($\pm 10^{\circ}$).

\paragraph{Prostate Cancer.}
The Prostate Cancer dataset~\citep{lawson2019prostate} consists of $5{,}182$ H\&E-stained ROIs extracted from $77$ whole-slide images, annotated into three Gleason grade categories (benign, Gleason 3, Gleason 4{+}5). We use a fixed patient-level split that combines training and validation for training ($4{,}145$ images) and evaluates on a held-out test set ($1{,}037$ images), preventing slide-level leakage across train and test. ROIs are resized to $224^2$, converted to 8-bit grayscale via the BT.601 luma transform, and augmented with random flips, $90^{\circ}$ rotations, and stain-neutral color jitter.

\paragraph{APTOS\,2019.}
APTOS\,2019~\citep{aptos2019} provides $3{,}662$ diabetic retinopathy fundus images labeled on a 5-point severity scale (0: no DR; 1: mild; 2: moderate; 3: severe; 4: proliferative). Since the Kaggle test labels are withheld, we re-split the public training set with a stratified random split (seed~$42$) that combines training and validation for training ($3{,}296$ images) and evaluates on the held-out test set ($366$ images), preserving the class distribution. Images are resized to $512^2$ using bilinear interpolation and converted to 8-bit grayscale via the BT.601 luma transform. We apply circular field-of-view cropping followed by random flips and rotation ($\pm 20^{\circ}$).

\paragraph{Segmentation datasets.}
\textbf{ISIC\,2018 (Task~1)}~\citep{isic2018} is used at $224^2$ for the main segmentation experiments, with the official challenge split combining training and validation for training ($2{,}698$ images) and the held-out test set ($1{,}002$ images). \textbf{CREMI}~\citep{cremi2016} (electron microscopy; neuron membrane segmentation) is used at two resolutions: $512^2$ crops for the main segmentation results in \Cref{tab:seg_accuracy}, and at the native $1250^2$ for the topology-aware loss experiments in \Cref{tab:topoloss}; both follow the same volume-level train/test split ($336$/$39$ slices). \textbf{Synapse}~\citep{landman2015miccai} (abdominal CT; 9-class multi-organ segmentation) is used at $224^2$ as 2D axial slices from the standard $18$/$12$ training/testing volume split following~\citep{chen2024transunet}, yielding $2{,}211$/$1{,}568$ slices. \textbf{Massachusetts Roads}~\citep{mnih2013} is used at native $1500^2$ for the topology-aware loss experiments, with the split defined by~\citet{lux2025topograph} ($1{,}122$/$49$). \textbf{Platelet}~\citep{guay2021platelet} (electron microscopy; 7-class organelles) is used at native $800^2$ following the same protocol ($50$/$24$).

\subsection{PH Algorithm Implementation Notes}
\label{app:ph-libraries}

All four PH algorithms that we compare compute zero- and one-dimensional cubical persistent homology ($H_0$ and $H_1$) on 2D images interpreted as sublevel-set filtrations of a scalar field, but they differ in their cubical complex conventions and internal algorithms.

\paragraph{T-construction vs.\ V-construction.}
GUDHI~\citep{maria2014gudhi} uses the T-construction, in which each pixel becomes a top-dimensional cube (a 2-cell) and lower-dimensional cells arise as shared boundaries. Cripser~\citep{kaji2020cubicalripser} and CMP~\citep{korkmaz2025cumperlay} use the V-construction, in which each pixel is a vertex (0-cell) and edges/faces arise from adjacency. The two conventions are dual under Alexander duality: $H_0$ under T-construction corresponds to $H_1$ under V-construction and vice versa on a 2D domain with appropriate boundary conditions~\citep{bleile2022persistent}. \cuph{} adopts the V-construction to match the convention used by Cripser and CMP, which simplifies direct comparison.

\paragraph{Equivalence of persistence diagrams.}
For integer-valued 2D inputs, all four algorithms produce identical persistence diagrams (up to the T/V convention swap) modulo the essential class treatment at infinity, which we verify empirically in \Cref{tab:correctness}. For floating-point inputs, \cuph{} and CMP require quantization to a finite number of filtration levels, while GUDHI and Cripser operate directly on floats; the quantization error is bounded by \Cref{cor:quantization-stability} and verified empirically in \Cref{tab:correctness}.

\paragraph{Parallelism.}
GUDHI is single-threaded per image but parallelized across a batch (\texttt{OMP\_NUM\_THREADS=16}). Cripser is similarly single-threaded per image and parallelized across the batch. CMP assigns one GPU thread per image, yielding batch-level but not within-image parallelism. \cuph{} exploits within-image parallelism via level-synchronous processing with up to $512$ threads per image, as detailed in \Cref{sec:gpu-impl}.

\subsection{Training Hyperparameters}
\label{app:hyperparameters}

All experiments use AdamW~\citep{loshchilov2019decoupled} (lr$\,{=}\,10^{-4}$, weight decay $0.01$) in fp32 with a cosine learning-rate schedule and linear warmup over the first $5\%$ of training steps. Classification follows PHG-Net's training setup~\citep{peng2024phg} at batch size $32$. Topology-aware loss experiments follow the Topograph protocol~\citep{lux2025topograph} (loss weights, image crops, optimization schedule). Segmentation experiments share the classification optimizer and schedule; per-dataset epoch counts and topology loss weights are released with the code.

\paragraph{TopoSwin-SP/MP configurations.}
We use the released configurations from~\citep{korkmaz2025cumperlay}: TopoSwin-SP uses single-parameter persistence with \texttt{filt\_rows}$=1$, and TopoSwin-MP uses multi-parameter persistence with \texttt{filt\_rows}$=8$. CMP runs with \texttt{OMP\_NUM\_THREADS=1} per the authors' documentation, since their implementation contends with the CUDA backend at higher thread counts.

\paragraph{Topology loss weights.}
For the topology-aware loss experiments (\Cref{tab:topoloss}), we use the loss weightings from~\citep{lux2025topograph}, each combined with a Dice $+$ cross-entropy pixel loss.

\subsection{Software and Hardware Stack}
\label{app:software}

All experiments run on a single NVIDIA H100 (80\,GB HBM3) GPU; CPU PH baselines run on $16$ threads of an AMD EPYC 7543 32-core processor (\texttt{OMP\_NUM\_THREADS=16}). The software stack includes CUDA, cuDNN, PyTorch (with \texttt{torch.compile} disabled for timing measurements to avoid JIT compilation overhead), GUDHI, Cripser (\texttt{CubicalRipser}), and CMP (the cubical PH backend in CuMPerLay). \cuph{} source code will be released upon acceptance.

\paragraph{Timing methodology.}
All GPU timings use \texttt{torch.cuda.synchronize()} before and after the measured region, with a $5$-iteration warmup before recording. Throughput is computed over full-epoch passes where indicated in the table caption and over fixed-size batches otherwise. CPU timings are wall-clock, measured with \texttt{time.perf\_counter()}. Accuracy and Dice metrics are reported as mean$_{\pm\mathrm{std}}$ over $5$ random seeds; throughput is the mean over $5$ runs.

\paragraph{Bit-depth and channel conversion for PH.}
PH operates on single-channel scalar fields. We process each dataset at its native bit depth to preserve intensity precision:

\begin{itemize}[nosep,leftmargin=*]
  \item \textbf{8-bit RGB} (ISIC\,2018 T1/T3, Prostate Cancer, APTOS\,2019, Roads): converted to single-channel 8-bit luminance via the ITU-R BT.601 transform, $Y = 0.299R + 0.587G + 0.114B$.
  \item \textbf{16-bit mammography} (CBIS-DDSM): raw 16-bit DICOM intensities, linearly rescaled to $[0, 65535]$ per image.
  \item \textbf{12-bit CT} (Synapse): Hounsfield units windowed to $[-125, 275]$, then linearly rescaled to the 12-bit range $[0, 4095]$.
  \item \textbf{8-bit EM} (CREMI): native uint8 grayscale, used directly.
  \item \textbf{16-bit EM} (Platelet): native uint16 grayscale, used directly.
\end{itemize}

The RGB backbone input (for classification and segmentation models that accept color) uses the original three-channel images without any conversion; only the PH pipeline consumes the single-channel intensities.

\begin{table}[t]
\caption{%
  PH throughput on $224^2$ ISIC and CBIS-DDSM at batch size $64$ ($H_0$ and $H_1$). GUDHI and Cripser run on $16$ CPU threads. Memory column reports peak GPU allocation.
}
\label{tab:ph_main}
\centering
\small
\begin{tabular}{@{}llrrr@{}}
\toprule
\textbf{Method} & \textbf{Device}
  & \textbf{ISIC (img/s)}
  & \textbf{CBIS (img/s)}
  & \textbf{Memory} \\
\midrule
GUDHI   & CPU & $97.5$    & $120.5$ & --- \\
Cripser & CPU & $458.5$   & $526.3$ & --- \\
CMP     & GPU & $56.2$    & $48.6$  & $994$\,MB \\
\textbf{\cuph{} (ours)}
        & \textbf{GPU}
        & $\mathbf{5{,}232}$
        & $\mathbf{5{,}023}$
        & $\mathbf{272}$\,MB \\
\bottomrule
\end{tabular}
\end{table}

\subsection{Full Classification Efficiency}
\label{app:full_classification}

\Cref{tab:efficiency-full} reports AUC, training throughput, and peak GPU memory for all classification methods, complementing \Cref{tab:cls_main}. Some entries use a smaller batch size where the default ($32$) exceeds GPU memory; these are marked $^{\dagger}$ (bs $=16$) and $^{\ddagger}$ (bs $=8$). On CBIS-DDSM, AUC is depressed for TopoSwin variants due to binary class imbalance, a regime where Acc and AUC diverge; following~\citep{korkmaz2025cumperlay} we report Acc as the primary metric in \Cref{tab:cls_main}.

\begin{table}[t]
\caption{%
  Full classification efficiency on a single H100. Each cell: AUC (\%) / throughput (img/s) / peak GPU memory (GB).
}
\label{tab:efficiency-full}
\centering
\footnotesize
\setlength{\tabcolsep}{3pt}
\begin{tabular}{@{}lcccc@{}}
\toprule
\textbf{Method} & \textbf{ISIC 2018 (T3)} & \textbf{CBIS-DDSM} & \textbf{Prostate} & \textbf{APTOS} \\
\midrule
SwinV2-B
  & $98.1_{\pm 0.5}$ / $206.7$ / $13.7$
  & $56.1_{\pm 8.5}$ / $206.1$ / $13.7$
  & $99.6_{\pm 0.1}$ / $208.5$ / $13.7$
  & $95.0_{\pm 0.4}$ / $55.2$ / $53.8$ \\
\midrule
PHG-Net
  & $98.3_{\pm 0.4}$ / $204.0$ / $14.0$
  & $71.7_{\pm 0.6}$ / $202.9$ / $14.0$
  & $99.7_{\pm 0.1}$ / $202.7$ / $14.0$
  & $94.8_{\pm 0.5}$ / $54.9$ / $54.2$ \\
PHG-Net (\cuph{})
  & $98.1_{\pm 0.1}$ / $183.9$ / $14.3$
  & $68.6_{\pm 1.7}$ / $183.4$ / $14.3$
  & $99.6_{\pm 0.1}$ / $183.7$ / $14.3$
  & $95.3_{\pm 0.3}$ / $49.5$ / $55.4$ \\
\midrule
TopoSwin-SP
  & $98.2_{\pm 0.2}$ / $16.4$ / $24.9$
  & $54.8_{\pm 6.4}$ / $16.8$ / $23.9$
  & $99.5_{\pm 0.4}$ / $17.9$ / $24.8$
  & $93.8_{\pm 0.2}$ / $2.9^{\dagger}$ / $54.9^{\dagger}$ \\
TopoSwin-SP (\cuph{})
  & $98.0_{\pm 0.2}$ / $116.9$ / $31.7$
  & $56.5_{\pm 4.8}$ / $111.1$ / $31.7$
  & $99.8_{\pm 0.1}$ / $118.5$ / $31.7$
  & $94.5_{\pm 0.6}$ / $28.8^{\dagger}$ / $71.1^{\dagger}$ \\
\midrule
TopoSwin-MP
  & $98.3_{\pm 0.1}$ / $12.1$ / $40.3$
  & $52.2_{\pm 2.4}$ / $12.4$ / $38.0$
  & $99.5_{\pm 0.3}$ / $11.9$ / $40.0$
  & $95.0_{\pm 0.4}$ / $0.6^{\ddagger}$ / $48.7^{\ddagger}$ \\
TopoSwin-MP (\cuph{})
  & $98.3_{\pm 0.2}$ / $52.6^{\dagger}$ / $53.4^{\dagger}$
  & $51.0_{\pm 2.1}$ / $52.6^{\dagger}$ / $53.4^{\dagger}$
  & $99.7_{\pm 0.2}$ / $53.3^{\dagger}$ / $53.5^{\dagger}$
  & $95.1_{\pm 0.5}$ / $10.0^{\ddagger}$ / $59.8^{\ddagger}$ \\
\bottomrule
\end{tabular}
\\[2pt]
\footnotesize
$^{\dagger}$bs $=16$.\quad $^{\ddagger}$bs $=8$.
\end{table}

\section{Correctness Verification}
\label{app:correctness}

This appendix empirically verifies \cuph{}'s correctness against Cripser~\citep{kaji2020cubicalripser} as a reference implementation, complementing \Cref{thm:equivalence} (integer correctness) and \Cref{cor:quantization-stability} (float quantization stability) in the main paper.

\paragraph{Experimental setup.}
We evaluate numerical agreement between \cuph{} and Cripser on $500$ randomly selected medical images per datatype at $224^2$ resolution (batch $64$). Three datatypes are examined: uint8 from ISIC\,2018 (T3), uint16 from BraTS~\citep{bakas2017advancing}, and float32 from BraTS with per-volume min-max normalization to $[0, 1]$. \Cref{tab:correctness} reports persistence-diagram agreement and throughput. Pair agreement is the fraction of $(\mathit{birth}, \mathit{death})$ pairs that match between \cuph{} and Cripser, reported separately for $H_0$ and $H_1$. We also report bottleneck distance (BD) and $2$-Wasserstein distance (WD$_2$) on the persistence diagrams; speedup is \cuph{} throughput divided by Cripser throughput on the same inputs.

\paragraph{Integer inputs.}
For uint8 and uint16 inputs, \cuph{} matches Cripser exactly: every $(\mathit{birth}, \mathit{death})$ pair value agrees, yielding $100\%$ pair agreement and BD $=$ WD$_2 = 0$. Both algorithms process the same finite set of integer filtration levels with identical merge-tree logic, so differences in scheduling do not affect the final pair values. \Cref{thm:equivalence} formally establishes this exact correspondence for finite-level filtrations.

\paragraph{Float32 inputs.}
For float32 inputs, \cuph{} discretizes the input into $L$ filtration levels before its bucket-parallel sweep, while Cripser operates on the native float representation (Cripser internally promotes float32 to float64). The discretization introduces a controlled approximation that satisfies the stability bound:
\[
d_B(\mathrm{Dgm}(f), \mathrm{Dgm}(f_L)) \leq \|f - f_L\|_\infty \leq \frac{\mathrm{range}(f)}{2L},
\]
where $f_L$ denotes the $L$-level discretization of $f$~\citep{cohen2007stability}. \Cref{tab:correctness} confirms this scaling empirically: BD decreases as $\mathcal{O}(1/L)$, reaching $2 \times 10^{-5}$ at $L = 2^{15}$, and WD$_2$ exhibits the same $\mathcal{O}(1/L)$ trend, decreasing from $4.4 \times 10^{-2}$ at $L = 2^{8}$ to $4.0 \times 10^{-4}$ at $L = 2^{15}$. The bound is tight in practice: empirical BD values match the theoretical $\mathrm{range}(f)/(2L)$ prediction within a small constant factor across the entire $L$ sweep. A residual ${\sim}0.2\%$ of pairs remain unmatched at $L = 2^{15}$, arising from tie-breaking when two pixels share the same quantized value; the resulting persistence diagrams remain within the BD bound and the unmatched pairs do not affect downstream applications, as confirmed by the matching classification (\Cref{tab:cls_main}) and segmentation (\Cref{tab:seg_accuracy}) results.

\paragraph{Throughput across $L$.}
\Cref{tab:correctness} also reports the speedup of \cuph{} over Cripser at each $L$. \cuph{} exceeds Cripser's throughput across the float32 $L$ sweep, with the advantage decreasing from $13.3\times$ at $L = 2^{8}$ to $1.8\times$ at $L = 2^{15}$ as the bucket-sweep cost grows linearly with $L$. We use $L = 2^{12}$ as the default for uint16 medical data, since the effective per-image entropy of typical 16-bit medical scans is well below $2^{12}$ distinct values.

\paragraph{Memory footprint.}
\Cref{tab:correctness_memory} reports memory usage for the correctness experiment. \cuph{}'s GPU allocation is $50$--$90\times$ smaller than Cripser's aggregated CPU resident set size in the $16$-worker configuration across all tested input types and $L$ values.

\begin{table}[t]
\caption{%
  Persistence-diagram agreement and throughput of \cuph{} vs.\ Cripser on $500$ images per datatype at $224^2$ (batch $64$). For float32 inputs, \cuph{} discretizes to $L$ filtration levels; Cripser uses the native-float representation. Pair agreement is the fraction of $(\mathit{birth}, \mathit{death})$ pairs that match, reported separately for $H_0$ and $H_1$. For integer inputs, $L$ is the natural number of distinct intensity values.
}
\label{tab:correctness}
\centering
\small
\setlength{\tabcolsep}{4pt}
\begin{tabular}{@{}l c cc cc c@{}}
\toprule
\textbf{Input} & $L$
  & \textbf{$H_0$ pairs} & \textbf{$H_1$ pairs}
  & \textbf{max BD} & \textbf{max WD$_2$}
  & \textbf{Speedup} \\
\midrule
uint8   & $2^{8}$  & $100\%$ & $100\%$ & $0$ & $0$ & $18.0\times$ \\
uint16  & $2^{12}$ & $100\%$ & $100\%$ & $0$ & $0$ & $5.4\times$  \\
\midrule
float32 & $2^{8}$  & $83.2\%$ & $85.5\%$ & $1.96\times10^{-3}$ & $4.4\times10^{-2}$ & $13.3\times$ \\
float32 & $2^{10}$ & $95.2\%$ & $95.9\%$ & $4.9\times10^{-4}$  & $1.2\times10^{-2}$ & $7.5\times$ \\
float32 & $2^{12}$ & $98.7\%$ & $99.0\%$ & $1.2\times10^{-4}$  & $3.0\times10^{-3}$ & $3.8\times$ \\
float32 & $2^{13}$ & $99.3\%$ & $99.5\%$ & $6.0\times10^{-5}$  & $1.4\times10^{-3}$ & $2.9\times$ \\
float32 & $2^{14}$ & $99.6\%$ & $99.8\%$ & $3.0\times10^{-5}$  & $7.0\times10^{-4}$ & $2.3\times$ \\
float32 & $2^{15}$ & $99.8\%$ & $99.8\%$ & $2.0\times10^{-5}$  & $4.0\times10^{-4}$ & $1.8\times$ \\
\bottomrule
\end{tabular}
\end{table}

\begin{table}[t]
\caption{%
  Memory footprint for the correctness experiment on $224^2$ images at batch $32$ ($16$-worker Cripser configuration). \cuph{} memory is GPU allocation; Cripser memory is CPU resident set size (RSS) aggregated across workers.
}
\label{tab:correctness_memory}
\centering
\small
\begin{tabular}{@{}llll@{}}
\toprule
\textbf{Input} & $L$
  & \textbf{\cuph{} GPU (batch $32$)}
  & \textbf{Cripser CPU RSS ($16$ workers, aggregate)} \\
\midrule
uint8   & $2^{8}$  & ${\sim}62$\,MB  & $5.64$\,GB ($352.7$\,MB $\times 16$) \\
uint16  & $2^{12}$ & ${\sim}65$\,MB  & $5.66$\,GB ($353.9$\,MB $\times 16$) \\
float32 & $2^{8}$  & ${\sim}62$\,MB  & $5.68$\,GB ($355.0$\,MB $\times 16$) \\
float32 & $2^{15}$ & ${\sim}109$\,MB & $5.68$\,GB \\
\bottomrule
\end{tabular}
\end{table}

\section{Ablation: Detailed Results}
\label{app:ablation}

\begin{table}[t]
\caption{%
  Ablation of \cuph{} on ISIC at $224^2$, batch $64$. Slowdown is the throughput ratio of full \cuph{} to the variant.
}
\label{tab:ablation}
\centering
\small
\begin{tabular}{@{}lrr@{}}
\toprule
\textbf{Variant} & \textbf{img/s} & \textbf{Slowdown} \\
\midrule
\cuph{} (full)                                          & $5{,}232$ & $1.0\times$ \\
\midrule
without Level Bucketing                                 & $1{,}139$ & $4.6\times$ \\
without Parallel Level Sweep                            & $249$     & $21.0\times$ \\
without Level Bucketing and Parallel Level Sweep        & $12.3$    & $425.4\times$ \\
\bottomrule
\end{tabular}
\end{table}

\Cref{tab:ablation} reports ablation results on uint8 ISIC at $L = 256$. This appendix extends those results across datatypes and engineering optimizations.

\subsection{Cross-datatype Trends}

\Cref{tab:ablation_dtypes} reports throughput and slowdown for each algorithmic ablation across uint16 and float32 inputs (uint8 results are in \Cref{tab:ablation}). Two patterns emerge.

\paragraph{Level Bucketing's contribution grows with $L$.}
At $L = 256$ (uint8), removing Level Bucketing yields a $4.6\times$ slowdown; at $L = 4096$ (uint16, float32), the slowdown rises to $6.4$--$6.7\times$. Larger $L$ requires the no-bucket variant to scan all $n$ pixels at more filtration levels, increasing the cost of removing the bucket structure proportionally.

\paragraph{Parallel Level Sweep's contribution shrinks with $L$.}
At $L = 256$, removing the parallel sweep yields a $21.0\times$ slowdown; at $L = 4096$, the slowdown drops to $7.8$--$8.3\times$. Larger $L$ means fewer pixels activate per level on average, leaving less within-level parallelism for the sweep to exploit.

The two contributions are complementary across the practical $L$ range: bucket sort dominates at high $L$, parallel sweep dominates at low $L$. Both are necessary for the full speedup.

\paragraph{Joint ablation supermultiplicativity intensifies at high $L$.}
At $L = 256$, removing both contributions yields a $425.4\times$ slowdown, $4.4\times$ above the multiplicative prediction of $\sim 97\times$. At $L = 4096$, the joint slowdown jumps to $1{,}878\times$ (uint16) and $1{,}779\times$ (float32), over $30\times$ above the product of individual ablations ($56\times$ and $50\times$). The gap between joint and individual slowdowns widens with $L$ because the no-bucket scan compounds across more levels when the parallel sweep is also removed, and a single thread cannot amortize the per-level scan cost.

\begin{table}[t]
\caption{%
  Cross-datatype ablation results at batch $64$. Throughput in img/s; slowdown is the throughput ratio of full \cuph{} to the variant; mean $\pm$ std over $5$ runs.
}
\label{tab:ablation_dtypes}
\centering
\small
\setlength{\tabcolsep}{4pt}
\begin{tabular}{@{}lrr@{}}
\toprule
\textbf{Variant} & \textbf{uint16 ($L = 2^{12}$)} & \textbf{float32 ($L = 2^{12}$)} \\
\midrule
\cuph{} (full)
  & $1{,}628 \pm 16$ ($1.0\times$)
  & $1{,}542 \pm 5$  ($1.0\times$) \\
\midrule
without Level Bucketing
  & $241 \pm 1$      ($6.7\times$)
  & $239 \pm 0.3$    ($6.4\times$) \\
without Parallel Level Sweep
  & $197 \pm 1$      ($8.3\times$)
  & $197 \pm 1$      ($7.8\times$) \\
without both
  & $0.87 \pm 0.0001$ ($1{,}878\times$)
  & $0.87 \pm 0.0004$ ($1{,}779\times$) \\
\bottomrule
\end{tabular}
\end{table}

\subsection{Engineering Ablations}

\Cref{tab:ablation_engineering} reports the slowdown when each engineering optimization is removed, across all three datatypes. Four observations:

\begin{itemize}
\item \textbf{Shared-memory histogram} contributes negligibly. Replacing the shared-memory accumulation with global-memory atomic adds yields essentially the same throughput ($\leq 1.01\times$) across all datatypes. The histogram phase is a small fraction of total runtime, and modern global atomics are fast enough on H100.

\item \textbf{GPU memory pool} contributes modestly. Replacing the persistent allocator with \texttt{cudaMalloc/Free} per call yields $1.08$--$1.19\times$ slowdown. The cost is amortized across the batch but not eliminated.

\item \textbf{Path halving in union-find} has no measurable benefit on these datasets. Removing path halving yields a slight throughput \emph{increase} ($0.97$--$0.99\times$ ratio of full to variant means the variant is $1$--$3\%$ faster). Component trees on BraTS and ISIC are shallow enough that path halving's amortized advantage is dominated by its constant-factor overhead in the find operation. We retain path halving in the production code because it provides asymptotic guarantees on adversarial inputs.

\item \textbf{Lock-free Component Tracking} is the largest engineering contributor. Replacing the lock-free \texttt{atomicCAS} merge with a global lock that serializes within-level merges yields a $19.8\times$ slowdown on uint8 ISIC. Activation-phase parallelism contributes only ${\sim}5\%$ of additional headroom over the no-parallel-sweep variant ($21.0\times$ slowdown), confirming that the lock-free design is what allows within-image parallelism to translate into throughput.
\end{itemize}

\paragraph{Thread-count saturation depends on $L$.}
\Cref{tab:ablation_threads} shows threads-per-block sensitivity. At $L = 256$ (uint8), reducing from $T = 512$ to $T = 64$ yields a $2.9\times$ slowdown. At $L = 4096$ (uint16, float32), the same reduction yields only $1.24$--$1.26\times$ slowdown. At higher $L$, fewer pixels activate per level, so within-level parallelism saturates with fewer threads, and the level loop itself dominates runtime regardless of thread count.

\begin{table}[t]
\caption{%
  Engineering ablation slowdowns at batch $64$. Slowdown is the throughput ratio of full \cuph{} to the variant; mean over $5$ runs.
}
\label{tab:ablation_engineering}
\centering
\small
\begin{tabular}{@{}lrrr@{}}
\toprule
\textbf{Removed optimization} & \textbf{uint8 ($L = 2^{8}$)} & \textbf{uint16 ($L = 2^{12}$)} & \textbf{float32 ($L = 2^{12}$)} \\
\midrule
shared-memory histogram & $1.00\times$ & $1.00\times$ & $1.00\times$ \\
GPU memory pool         & $1.19\times$ & $1.09\times$ & $1.08\times$ \\
path halving            & $0.99\times$ & $0.97\times$ & $0.98\times$ \\
lock-free Component Tracking & $19.8\times$ & --- & --- \\
\bottomrule
\end{tabular}
\end{table}

\begin{table}[t]
\caption{%
  Threads-per-block sensitivity at batch $64$. Slowdown is the throughput ratio of $T = 512$ (default) to the variant; mean over $5$ runs.
}
\label{tab:ablation_threads}
\centering
\small
\begin{tabular}{@{}rrrr@{}}
\toprule
\textbf{Threads / block} & \textbf{uint8 ($L = 2^{8}$)} & \textbf{uint16 ($L = 2^{12}$)} & \textbf{float32 ($L = 2^{12}$)} \\
\midrule
$512$ (default) & $1.0\times$  & $1.0\times$  & $1.0\times$  \\
$256$           & $0.99\times$ & $1.00\times$ & $1.00\times$ \\
$128$           & $1.61\times$ & $1.09\times$ & $1.09\times$ \\
$64$            & $2.90\times$ & $1.24\times$ & $1.26\times$ \\
\bottomrule
\end{tabular}
\end{table}

\subsection{Correctness Validation}

All ablation variants produce persistence diagrams that match \cuph{} (full) bit-exactly on integer inputs. Validation across $11$ variants, $3$ datatypes, $5$ test images, and both $H_0$ and $H_1$ ($330$ set-equality checks total) confirms exact correspondence. The path-halving ablation, despite removing the path-shortening optimization, preserves correctness: the union-find merge logic is independent of compression strategy.

\subsection{Implementation Notes}

\paragraph{Without Level Bucketing.}
The histogram, prefix-sum, and scatter kernels are skipped entirely. The main kernel iterates all $n$ pixels at every level with a per-pixel level-equality check (\texttt{round(image[pixel\_idx]) == level}), activating and connecting only those pixels whose value matches the current level.

\paragraph{Without Parallel Level Sweep.}
Both the activate and connect phases of \texttt{cuph\_process\_levels\_kernel} are wrapped in \texttt{if (threadIdx.x == 0) \{ ... \}}. Other threads in the block remain idle. \texttt{\_\_syncthreads()} barriers remain in place to preserve block well-formedness; per-phase synchronization inside the divergent branch would invoke undefined behavior. The atomic-CAS retry loop in \texttt{cg\_merge\_components} succeeds on the first attempt under single-thread execution, so the backoff path is never exercised.

\paragraph{Without both.}
Both modifications are combined: histogram/prefix-sum/scatter kernels are skipped, and the main kernel processes all $n$ pixels at every level using only thread $0$.

\paragraph{Without lock-free Component Tracking.}
The activation phase remains parallel across $512$ threads, but the connect phase serializes all merges within each filtration level: the merge operations are wrapped in \texttt{if (threadIdx.x == 0)} while the activation phase runs in parallel. Path halving in $\textsc{Find}$ is retained to isolate the contribution of the lock-free \texttt{atomicCAS} primitive specifically.

\end{document}